\documentclass{article}

\usepackage[preprint]{neurips_2026}

\usepackage[utf8]{inputenc} 
\usepackage[T1]{fontenc}    
\usepackage{hyperref}       
\usepackage{url}            
\usepackage{booktabs}       
\usepackage{amsfonts}       
\usepackage{nicefrac}       
\usepackage{microtype}      
\usepackage{xcolor}         
\usepackage{soul}
\usepackage{algorithm}
\usepackage{algorithmic}
\usepackage{amsmath}
\usepackage[dvipsnames]{xcolor}
\usepackage{graphicx}
\usepackage{subcaption}
\usepackage{bbm}
\usepackage{subcaption} 
\usepackage{fontawesome5}

\newcommand{\abs}[1]{\left\lvert #1 \right\rvert}

\usepackage{standalone}
\usepackage{tikz}
\usetikzlibrary{shapes.geometric, arrows.meta, positioning, fit, calc, backgrounds}
\usepackage{amsmath, amssymb}
\usepackage{wrapfig}
\usepackage[font=small,labelfont=bf]{caption}
\usepackage[table]{xcolor}
\usepackage{booktabs}
\usepackage{graphicx}
\usepackage{amsfonts}
\usepackage{multirow}
\title{Contextual Plackett–Luce: An Efficient Neural Model for Probabilistic Sequence Selection under Ambiguity}

\author{%
  Noam Mizrachi \thanks{Equal contribution} \\
  Mobileye \\
  \texttt{Noam.Mizrachi@mobileye.com} \\
  \And
  Nadav Har-Tuv \footnotemark[1] \\
  Mobileye \\
  \texttt{Nadav.Har-Tuv@mobileye.com} \\
  \And
  Shai Shalev-Shwartz \\
  Mobileye\\
  \texttt{Shai.Shalev-Shwartz@mobileye.com} \\
}

\begin{document}

\maketitle

\begin{abstract}
\label{sec:abstract}
Selecting a coherent sequence or subset of elements is a fundamental problem in structured prediction, arising in tasks such as detection, trajectory forecasting, and representative subset selection. In many such settings, the target is inherently ambiguous: each input admits multiple valid outputs, while supervision provides only a single sampled instance. This induces a mismatch between the underlying multi-modal target distribution and the observed training signal.
We propose Contextual Plackett–Luce (CPL), a structured probabilistic model for sequence selection that extends the classical Plackett–Luce model to a context-dependent setting following an Ising-style parameterization with unary and pairwise interaction terms. CPL can be viewed as a hybrid between fully autoregressive prediction and parallel sequence selection: autoregressive models effectively capture uncertainty but are computationally expensive on modern parallel hardware such as GPUs, while parallel methods are efficient but struggle to represent multi-modal dependencies. CPL combines the strengths of both by constructing the parameters of a probabilistic selection model in a fully parallel manner, followed by a lightweight autoregressive selection process in which each step applies incremental updates to contextual logits. This decoupling of parallel scoring and sequential selection enables efficient computation without sacrificing expressivity.
We evaluate CPL on two structured selection tasks: multi-modal path prediction and representative subset selection. CPL achieves improved structural consistency and robustness under ambiguous supervision compared to strong parallel baselines.
\end{abstract}

\begin{center}
\noindent \faGithub\ Source code: \href{https://github.com/cplPaper2026/CPL_NeurIPS_2026}{https://github.com/cplPaper2026/CPL\_NeurIPS\_2026}.
\vspace{-0.5em}
\end{center}
\vspace{-0.5em}

\section{Introduction}
\label{sec:introduction}

Structured prediction problems often require selecting a coherent subset or sequence of elements from a large candidate space~\citep{carion2020detr,vinyals2015pointer,kulesza2012dpp}. In such settings, the correctness of one decision depends on the configuration of the others: a useful output is not merely a collection of high-scoring elements, but a structured object whose parts must be mutually consistent.

A central difficulty arises when the target is inherently ambiguous. The same input may admit several valid structured outputs, while training supervision provides only a single observed instance. This creates a mismatch between the multi-modal structure of the task and the single realization seen during learning~\citep{rhinehart2018r2p2,chai2020multipath}.

\paragraph{Running example: path prediction.}

Consider predicting a future driving path from road geometry. 
This problem is central in autonomous driving, where the goal is to plan the ego trajectory and to predict the future behavior of other agents in the scene~\citep{caesar2020nuscenes,phanminh2020covernet,chai2020multipath}. 
In practice, supervision for this task is typically constructed from recorded trajectories, using the future path that an agent actually followed. 
However, this reflects only a single realized outcome, even though multiple valid alternatives may have been available at that moment. This mismatch between multiple possible futures and a single observed trajectory introduces inherent ambiguity into the learning problem.

\begin{wrapfigure}{r}{0.4\textwidth}
    \centering
    \includegraphics[
        width=0.48\textwidth,
        trim={70 80 0 90},
        clip
    ]{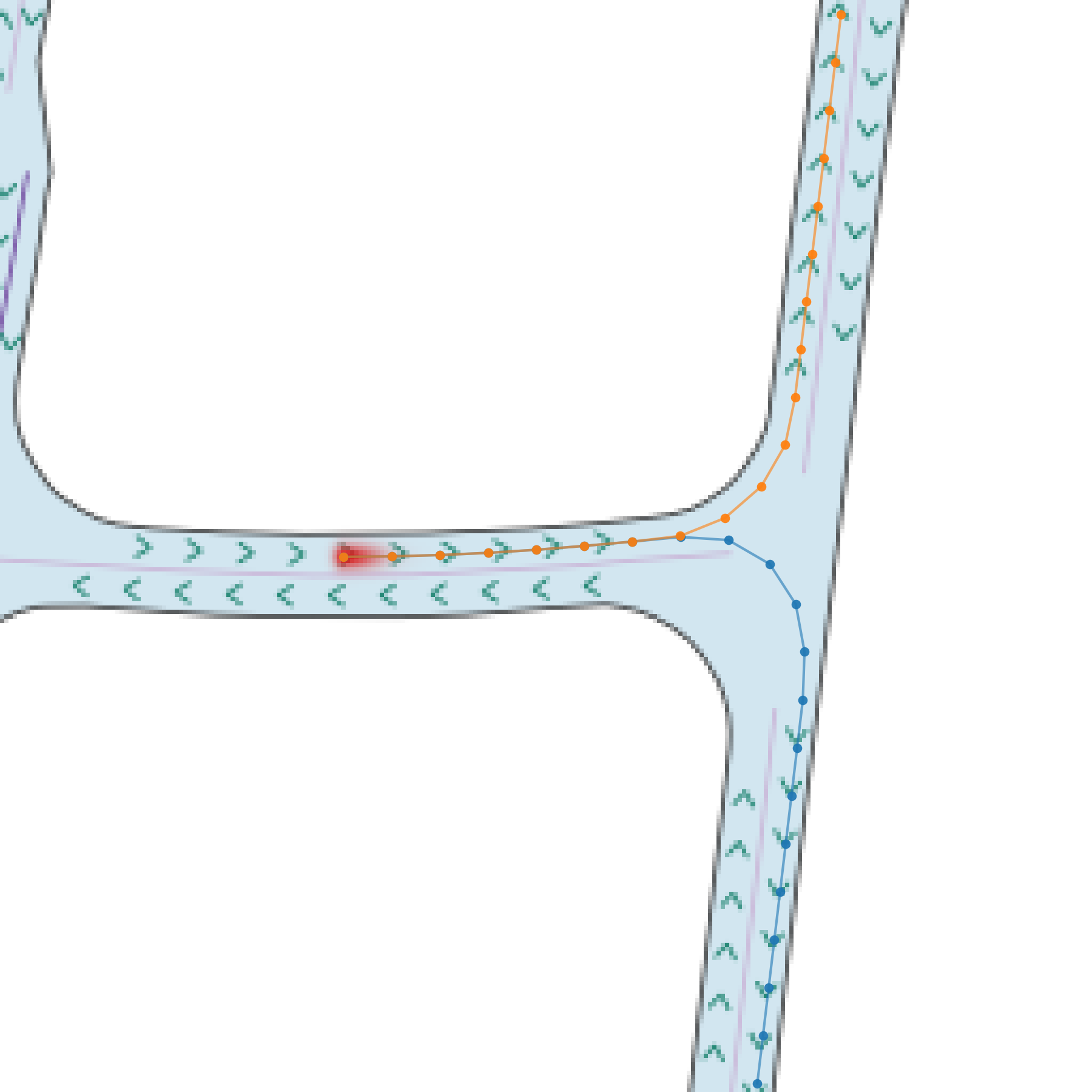}
    \textbf{(a)}

    \vspace{0.8em}

    \includegraphics[
        width=0.48\textwidth,
        trim={70 80 0 90},
        clip
    ]{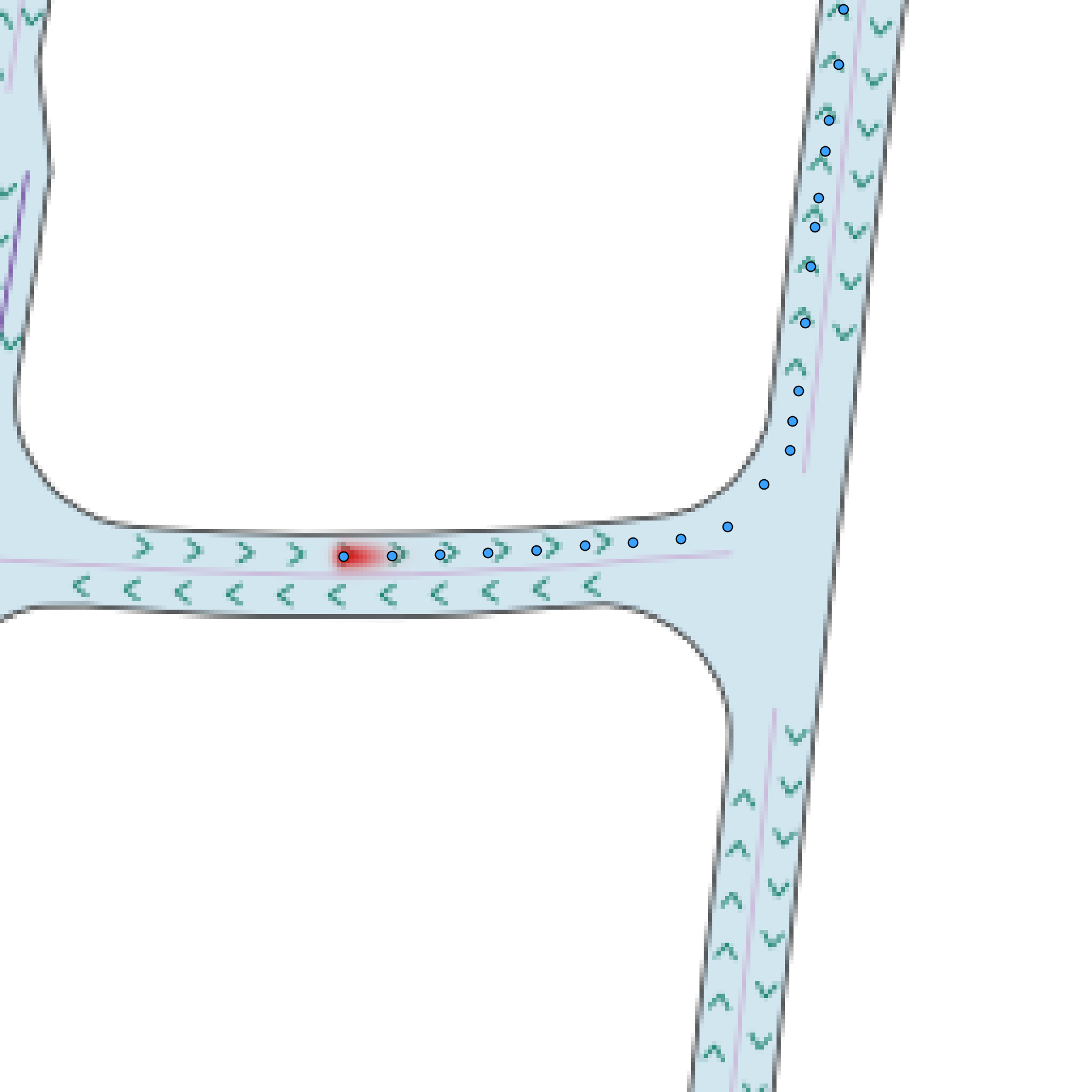}
    \textbf{(b)}
    
    \caption{Path prediction under ambiguity. (a) Ground-truth trajectories in a scene with multiple valid continuations. (b) CPL prediction selecting a single consistent branch.}
    \label{fig:path_example}
\end{wrapfigure}

As illustrated in Figure~\ref{fig:path_example}, junctions and splits may admit multiple valid continuations. A training set can therefore contain different valid choices for similar inputs, for example a left continuation in one instance and a right continuation in another. This exposes the model to multiple plausible but mutually exclusive outcomes, while never observing the full set of possibilities for any given scene.

At inference time, the desired output is a single coherent path: the model should commit to one branch and continue along it, rather than merging incompatible alternatives.

This example exposes the core challenge studied in this paper. 
Independent scoring methods are efficient, but they do not model the interactions needed to suppress redundant or incompatible choices, and can therefore produce fragmented outputs. 
Matching-based set predictors introduce global alignment, but under single-sample supervision they may favor intermediate or hybrid configurations that partially match multiple valid modes without corresponding to any valid output~\citep{bishop1994mixture,rupprecht2017learning}. 
Appendix~\ref{app:ambiguity_example} provides a minimal illustrative example of this failure mode.
Fully autoregressive models naturally resolve ambiguity by committing one decision at a time, but their sequential computation can be expensive for dense candidate spaces.

\paragraph{Our approach.}
We introduce Contextual Plackett--Luce (CPL), a structured probabilistic framework for subset and sequence selection under ambiguous supervision. CPL builds on the normalized sequential-choice structure of Plackett--Luce models~\citep{luce1959choice,plackett1975analysis,maystre2015plackettluce} while extending it with context-dependent utilities: each selected element updates the logits of the remaining candidates through learned pairwise interactions. This enables history-dependent decision making without recomputing representations at each step, as the underlying features are computed once and reused throughout the selection process. CPL supports both ordered sequences and unordered subsets under a common formulation.

We evaluate CPL in two complementary regimes: ordered path prediction and unordered representative subset selection.
Multi-modal path prediction requires selecting a single coherent trajectory despite multiple valid continuations. 
Representative subset selection requires choosing a subset of representative elements that captures the underlying structure of the data while avoiding redundant or conflicting selections under ambiguous supervision.

Our main contributions are:
\begin{itemize}\itemsep0pt
    \item a contextual Plackett--Luce model that augments unary scores with learned pairwise interactions, enabling history-dependent selection and explicit modeling of compatibility between elements;
    
    \item a permutation-invariant training objective for learning from unordered and ambiguous supervision, allowing the model to recover consistent structures from single sampled targets;
    
    \item an efficient decoding procedure in which each selection step reduces to a lightweight logit update using precomputed interactions, avoiding full autoregressive recomputation;
    
    \item empirical results on path prediction and representative subset selection demonstrating improved structural consistency and robustness under ambiguous supervision.
\end{itemize}
\section{Analysis of modeling trade-offs and constraints}
\label{sec:method}

To motivate our design, we first examine three standard paradigms for subset selection, highlighting their modeling limitations under ambiguous supervision and their computational properties.

\subsection{Preliminaries and notation}

We define the candidate set as $\mathcal{X} = [k]$, where each index $i \in \mathcal{X}$ corresponds to an embedding vector $q_i \in \mathbb{R}^d$. The goal is to select a subset $S \subseteq \mathcal{X}$ that forms a valid structured output.

We denote by $k$ the number of candidate elements, by $d$ the embedding dimension, and by $L$ the number of layers in the backbone network used to compute representations. The quantity $|S|$ denotes the size of the selected subset.

We distinguish between sequential and parallel computation, where $A \times B$ denotes a process with $A$ sequential steps and $B$ parallel operations per step.

\subsection{Existing paradigms and limitations}
\label{sec:method:baselines}


\paragraph{Independent scoring}

Independent scoring evaluates each candidate element in isolation, assigning a scalar utility that reflects its individual suitability.
Formally, each candidate $q_i$ is assigned a score $\theta_i$ and selected via thresholding:
\[
\theta_i = f_\theta(q_i), \quad S = \{ i \mid \theta_i > \tau \}.
\]

This formulation lacks a mechanism to coordinate selections, which can lead to over- or under-selection. In the running path prediction example, multiple branches at a junction may exceed the threshold, resulting in incompatible continuations being selected together, while in other cases valid segments may be suppressed.
We refer to this failure mode as \textit{fragmentation}: the output decomposes into disconnected components that are locally plausible but do not form a coherent solution.

The computation is fully parallel over candidates, requiring a single forward pass with complexity:
$
O(L \times d^2 k)
$.

\paragraph{Matching-based set prediction}

Matching-based methods shift supervision from individual elements to sets, training the model to predict a full configuration that aligns with the ground truth under a permutation-invariant matching objective. In practice, elements are first contextualized using a Transformer:
\[
(\theta_1, \dots, \theta_k) = \mathrm{Transformer}(q_1, \dots, q_k),
\]
so each prediction depends on the full set.

This enables set-level reasoning and global interactions, but remains limited under ambiguous supervision.
When multiple valid outputs exist, training data often provides only one realization per instance.
For similar inputs, the model can therefore observe conflicting targets, and minimizing an average prediction loss is known to favor conditional averages rather than committing to a single mode~\citep{bishop1994mixture,rupprecht2017learning}.
In structured tasks, this \textit{mode averaging} can produce intermediate configurations that do not correspond to any valid output.

The computation is dominated by full set-level interaction through self-attention, leading to an inference cost of
$
O(L \times (d^2 k + k^2 d)).
$


\paragraph{Autoregressive sequence selection}

Autoregressive models generate the subset sequentially~\cite{vinyals2015pointer}, where each decision is conditioned on previously selected elements:
\[
P(S) = \prod_{t} P(i_t \mid i_{<t}, \mathcal{X}), \quad i_t \in [k]
\]

This formulation naturally resolves ambiguity through sequential commitment: the model selects a single element at each step, and all subsequent decisions are conditioned on that choice. As a result, it produces a single coherent structure rather than blending multiple alternatives.

In the path prediction example, the model constructs the trajectory step by step. At a junction, it selects one branch (e.g., left or right), and all subsequent points follow that branch, ensuring consistency.

However, this expressiveness comes at a computational cost. Each step depends on the full history and typically requires recomputing context-dependent representations. Consequently, inference complexity grows with the subset size.

More precisely, beyond the shared backbone cost, sequential decoding introduces step-wise computation over the subset size. The overall inference complexity is:
\[
O(L \times (d^2 k + k^2 d) + |S| \times L \times (d^2 + k d))
\]
where the first term corresponds to the backbone encoding and the second term to autoregressive decoding.
Thus, autoregressive models handle ambiguity effectively, but are limited by their sequential computation.


\section{Contextual Plackett--Luce}
\label{sec:method:cpl}

We define CPL as a probabilistic selection module built on top of precomputed element representations, following the sequential-choice structure of Plackett--Luce models~\cite{luce1959choice,plackett1975analysis,maystre2015plackettluce}.

The model constructs all parameters governing sequential decisions in parallel over the full candidate set, encoding how previously selected elements influence remaining ones, and follows an autoregressive process that updates contextual logits using these precomputed interactions while avoiding recomputation of representations.


CPL parameterizes the next-element distribution using unary relevance scores and pairwise interactions:
\[
\theta_i = f_\theta(q_i),
\qquad
W_{ij} = f_W(q_i,q_j).
\]

Given a partially constructed subset $S_t$, candidate logits and next-element probabilities are:
\[
\ell_j(S_t) = \theta_j + \sum_{i\in S_t} W_{ji}.
\qquad
P(j \mid S_t) =
\frac{\exp(\ell_j(S_t))}
{\sum_{k \notin S_t} \exp(\ell_k(S_t))}.
\]

Thus, CPL models subset construction as a sequence of conditional decisions, where each choice reshapes the preference landscape over the remaining candidates.


At inference time, the subset is constructed via greedy decoding over the evolving distribution:
\[
j^* = \arg\max_{j \notin S_t} \ell_j(S_t)
\]
with the sequence terminating when the model selects a learned End-Of-Sequence (\texttt{EOS}) token.

Despite its autoregressive nature, CPL avoids recomputing deep network representations at each step. 
Because the interactions $W$ are pre-computed, the effect of adding a new element $j^*$ reduces to a simple vector accumulation:
\[
\ell(S_{t+1}) = \ell(S_t) + W_{:,j^*}.
\]
This isolates the sequential bottleneck to a lightweight addition operation. The overall runtime complexity is:
\[
O(L \times (d^2 k + k^2 d) + |S| \times k)\]
yielding the structured dependence of an autoregressive model at a fraction of the computational cost.

\subsection{Training}
\label{sec:method:cpl_training}

CPL is trained using a teacher-forcing scheme~\cite{williams1989teacher} and can be interpreted from two complementary perspectives corresponding to ordered and unordered formulations of subset selection.

In the ordered view, CPL is interpreted as a next-element prediction model. Given a partially constructed subset $S_t$, the model predicts the next element in the sequence using teacher forcing, where the ground-truth prefix is provided during training. In this setting, there is a single valid target $j^\star$, corresponding to the next element in the ground-truth ordering. The model is trained to maximize its probability under a context-dependent distribution:
\[
P_\text{model}(j^\star \mid S_t) = \mathrm{softmax}(\theta + W\,\mathbf{1}_{S_t} - \infty\,\mathbf{1}_{S_t})_{j^\star}.
\]

Here, $\mathbf{1}_{S_t} \in \{0,1\}^k$ is an indicator vector denoting previously selected elements. The $-\infty$ masking term excludes already selected elements from the candidate set in a vectorized manner, enabling efficient batched computation across different partial states while preserving valid sequence construction.

In the unordered setting, no canonical ordering is assumed. We therefore apply the same teacher-forcing mechanism over randomly sampled partial subsets $S \subseteq S^\star$, which simulate different valid construction prefixes. The remaining valid candidates are defined as $R = S^\star \setminus S$, with $R = \{\texttt{EOS}\}$ when $S = S^\star$. In contrast to the ordered case, all elements in $R$ are treated as equally valid next choices, inducing a uniform target distribution over valid continuations.

The training objective maximizes the expected likelihood over these sampled partial contexts:
\[
\mathcal{L}_{\mathrm{CPL}}
=
-\mathbb{E}_{S \subseteq S^\star}
\left[
\frac{1}{|R|}
\sum_{j^\star \in R}
\log P_{\mathrm{model}}(j^\star \mid S)
\right].
\]

This formulation unifies ordered next-step prediction and unordered set construction within the same probabilistic model. In both cases, CPL is trained in a teacher-forcing manner over partial structures, while differing only in whether supervision specifies a single next element or a uniform distribution over valid continuations.

\subsection{Behavior under ambiguity}
\label{sec:method:cpl_ambiguity}

Under ambiguous supervision, the interaction matrix provides a mechanism for progressive commitment.
After selecting an element from one plausible mode, negative interactions can suppress incompatible alternatives, while positive interactions can promote candidates that are coherent with the partial output.
For example, selecting a point on one branch of a road junction can reduce the logits of candidates on competing branches and favor later points along the same branch.
Thus, CPL encodes mode compatibility in precomputed pairwise terms and applies it through lightweight incremental logit updates, rather than full autoregressive recomputation.
Additional qualitative examples and visualizations of the learned interaction matrix are provided in Appendix~\ref{appendix:greedy_decoding}.
\section{Experiments}
\label{sec:experiments}

We evaluate CPL in two complementary structured-selection regimes.
The first is an ordered setting, where the output is a path and structural consistency requires committing to one continuation over time.
The second is an unordered setting, where the output is a representative subset and structural consistency requires covering latent groups without redundant selections.
Together, these experiments test whether history-dependent selection with lightweight contextual updates can resolve ambiguous single-sample supervision across both ordered and unordered outputs.

\subsection{Ordered structured selection: path prediction}
\label{sec:path-prediction}

We evaluate CPL on a multi-modal path-prediction benchmark designed to isolate the effect of ambiguous supervision.
Given a BEV road map and an ego pose, the model must predict one valid driving path.
The challenge is not to enumerate every possible continuation, but to commit to a single geometrically valid path when several futures are consistent with the scene.

\subsubsection{Setup}
\label{sec:path-setup}

\textbf{Task.}
The input is a BEV RGB map encoding road layout, lane directions, and the ego pose (Figure~\ref{fig:path_example}). The map is discretized into a coarse decision grid obtained by downsampling the full-resolution BEV with factor $D$, where each grid cell corresponds to a $D \times D$ region in pixel space.

Each cell $(i,j)$ is associated with a feature embedding and predicts a sub-cell offset $(\delta_x, \delta_y)$ that refines its position within the cell. Full-resolution coordinates are reconstructed as:
\begin{equation}
\label{equation:grid_offset}
(x,y)_{(i,j)} = (jD + \delta_x, iD + \delta_y).
\end{equation}

The path is modeled as an ordered subset of grid cells, selected from the grid together with an End-of-Sequence (EOS) token.

\textbf{Data.}
We construct the benchmark from real-world maps provided by nuScenes~\cite{caesar2020nuscenes}. Each sample is generated from a local road topology around the ego position and contains multiple valid ground-truth paths, including intersections, forks, and curved continuations. During training, only one valid path is sampled as supervision, so the model observes a single realization of a multi-modal target.

\textbf{Evaluation metrics.}
All distances are reported in full-resolution pixels and reduced as a $\min$ over the $K$ valid ground-truth paths per scene. We report $\min$-$\mathrm{ADE}$, which measures average displacement from the closest valid path, and $\min$-$\mathrm{HD}$, which captures the worst deviation from that path~\cite{huttenlocher1993hausdorff}.

We additionally report the off-road rate, defined as the fraction of predicted points lying outside the drivable area. To analyze performance under different ambiguity levels, we further stratify $\min$-$\mathrm{ADE}$ by the number of valid paths $n_{\mathrm{paths}}$ per scene.

Runtime is measured on the full validation set using 10 independent runs. For each run, we compute the mean per-example inference time over the dataset. We then report the mean and standard deviation across these 10 run-level averages, on a single NVIDIA A100 GPU.

\textbf{Implementation details.}
All methods use an ImageNet-pretrained ResNet-18 backbone~\cite{russakovsky2015imagenet,he2016resnet} followed by a Transformer encoder~\cite{vaswani2017attention}, producing per-grid-cell feature embeddings. The offset head follows the formulation in the Task paragraph, predicting sub-cell refinements for selected grid cells~(see Eq.~\ref{equation:grid_offset}). The compared methods differ only in the selection mechanism over grid cells. Additional training and runtime details are provided in Appendix~\ref{app:path-prediction-details}.

\subsubsection{Baselines}
\label{sec:path-baselines}

We compare CPL against four baselines covering the main structured selection paradigms in Section~\ref{sec:method}.

\textbf{Grid thresholding.}
A fully parallel baseline that predicts per-cell path occupancy using binary cross entropy, together with an $\ell_1$ offset regression on positive cells. At inference, cells are selected independently via thresholding, without modeling interactions between them.

\textbf{Hungarian set prediction.}
A set-based baseline trained with bipartite matching between predicted grid cells and ground-truth path points~\cite{kuhn1955hungarian,carion2020detr}. This removes dependence on a fixed assignment during training, but inference remains parallel, relying on independent per-cell scores.

\textbf{Multi-hypothesis prediction.}
A parallel model that predicts multiple candidate paths and selects the best-scoring hypothesis~\cite{chai2020multipath,phanminh2020covernet}. While it represents multiple modes, each hypothesis is generated independently without explicit sequential dependency.

\textbf{Autoregressive pointer network.}
A sequential model that generates path cells one at a time using a Transformer decoder~\cite{vinyals2015pointer,vaswani2017attention}. It captures history-dependent structure but requires full autoregressive decoding, leading to higher inference cost compared to parallel methods.

\begin{table}[h]
\centering
\caption{Ordered structured selection on the path-prediction benchmark
(validation split).
Distances are reported in full-resolution pixels and reduced as a $\min$ over
the valid ground-truth modes per scene.
Runtime is reported as mean ± standard deviation. Bold denotes the best result in each column.}
\label{tab:path-prediction:main}
\small
\begin{tabular}{lcccc}
\toprule
\textbf{Method} & $\min$-ADE $\downarrow$ & $\min$-HD $\downarrow$ & Off-road (\%) $\downarrow$ & Runtime (ms) $\downarrow$ \\
\midrule
Hungarian set prediction        & 8.20 & 29.14 & 10.91 & \textbf{2.79} $\pm$ 0.05 \\
Grid thresholding               & 5.84 & 35.76 & 0.52  & 2.80 $\pm$ 0.07 \\
Multi-hypothesis prediction     & 6.83 & 38.63 & \textbf{0.42} & 3.15 $\pm$ 0.09 \\
Autoregressive pointer          & 2.68 & 10.03 & 0.80  & 32.91 $\pm$ 0.14 \\
\midrule
\textbf{CPL}                    & \textbf{2.35} & \textbf{9.92} & 0.83 & 6.07 $\pm$ 0.14 \\
\bottomrule
\end{tabular}
\end{table}

\subsubsection{Results}
\label{sec:path-results}

Table~\ref{tab:path-prediction:main} reports global validation performance.
CPL achieves the best distance-based metrics while remaining substantially faster than the fully autoregressive pointer model.
The AR baseline confirms that sequential conditioning is effective for resolving ambiguous futures, but its decoding loop is costly.
CPL retains this commitment mechanism through lightweight contextual logit updates.

The off-road metric indicates that staying within drivable regions alone is insufficient to resolve ambiguity: most models correctly remain on-road, yet parallel methods can still combine incompatible continuations within the same prediction.
Hungarian set prediction improves supervision at the set level, but without history-dependent decoding it is still prone to mode averaging, producing outputs that lie between valid branches rather than committing to a single consistent path (see Appendix~\ref{path_prediction_viz}).


\textbf{Robustness to supervision ambiguity.}
Table~\ref{tab:path-stratified} stratifies $\min$-$\mathrm{ADE}$ by the number of valid ground-truth paths.
Parallel baselines degrade as the number of valid modes increases, whereas CPL and AR remain more stable, showing the benefit of conditioning on previous selections.
AR is slightly stronger in the most ambiguous cases, while CPL offers a better accuracy--runtime tradeoff.

\begin{table}[h]
\centering
\caption{$\min$-ADE stratified by the number of valid ground-truth modes
$n_{\mathrm{paths}}$.}
\label{tab:path-stratified}
\small
\begin{tabular}{lcccc}
\toprule
\textbf{Method} & $n_{=1}$ $\downarrow$ & $n_{=2}$ $\downarrow$ & $n_{=3}$ $\downarrow$ & $n_{\geq4}$ $\downarrow$ \\
\midrule
Hungarian set prediction        & 5.38 & 13.78 & 11.63 & 13.95 \\
Grid thresholding               & 4.86 & 6.48  & 9.11  & 12.54 \\
Multi-hypothesis prediction     & 6.13 & 6.96  & 9.05  & 13.61 \\
Autoregressive pointer          & 2.58 & 2.50  & \textbf{3.90} & \textbf{3.70} \\
\midrule
\textbf{CPL}                    & \textbf{2.08} & \textbf{2.25} & 4.12 & 4.52 \\
\bottomrule
\end{tabular}
\end{table}

Overall, the results show that robust path prediction under single-sample ambiguous supervision requires branch commitment and suppression of incompatible alternatives.
CPL provides this behavior without full autoregressive recomputation.
\subsection{Unordered structured selection: representative subset selection}
\label{sec:representative-subset-selection}

We evaluate CPL on representative subset selection as the unordered counterpart to the path-prediction experiment.
Each input is a set of image embeddings with latent semantic clusters, and the desired output is a compact subset that selects one representative from each cluster.
Supervision is incomplete: at every training draw, the target contains only one randomly sampled element per cluster, although many elements in that cluster would be equally valid.
Thus, the model never observes cluster boundaries or all valid representatives, but must still infer the invariant structure: cover all clusters, avoid duplicates, and stop at the right subset size.

\subsubsection{Setup}
\label{sec:subset-setup}

\textbf{Task.}
Let $X=\{x_i \in \mathbb{R}^d\}_{i=1}^N$ be an unordered set whose elements belong to latent clusters $\mathcal{C}(X)=\{C_c\}_{c=1}^k$.
The model observes only $X$ and predicts a subset $S \subseteq X$.
A prediction is structurally correct when every latent cluster is represented and no cluster is selected repeatedly.
At inference time, neither the cluster identities nor $k$ are given.

\textbf{Data.}
We build set instances from CIFAR-100 image classes~\cite{krizhevsky2009cifar}, using frozen ImageNet-pretrained ResNet-18 features~\cite{russakovsky2015imagenet,he2016resnet}.
Each instance samples a variable number of classes and images per class, yielding heterogeneous cluster sizes and set cardinalities.
For every draw of a training instance, the positive set is resampled by choosing one current element from each present class.
This prevents the learner from treating a particular exemplar as the target and makes the ambiguity explicit.
Reproducibility details are provided in Appendix~\ref{appendix:cifar_clustering_details}.

\textbf{Evaluation metrics.}
Since exact target elements are arbitrary, evaluation is performed at the cluster level.
For a predicted subset $S$, we report
\[
\mathrm{CluRec}
=
\frac{1}{k}
\sum_{c=1}^{k}
\mathbb{I}\!\left\{|S \cap C_c| > 0\right\},
\qquad
\mathrm{CluPrec}
=
\frac{1}{|S|}
\sum_{c=1}^{k}
\mathbb{I}\!\left\{|S \cap C_c| > 0\right\}.
\]
with precision defined as zero when $S$ is empty.
We also report $\text{CluF1}$ and $\text{CardErr}=\abs{|S|-k}$.
Together, these measure the three desired properties: coverage, low redundancy, and correct termination.

Runtime is measured on 1,000 random samples over 10 independent runs. For each run, we compute the mean per-example inference time, and report the mean and standard deviation across the 10 run-level averages, on a single NVIDIA A100 GPU.

\subsubsection{Baselines}
\label{sec:subset-baselines}

We compare CPL against baselines representing the same selection paradigms in the unordered setting.

\textbf{BCE thresholding.}
A parallel unary baseline trained with binary cross-entropy on sampled representatives. At inference, elements are selected by score thresholding.

\textbf{Hungarian set prediction.}
A matching-based baseline trained using bipartite assignment between predictions and ground-truth representatives~\cite{kuhn1955hungarian,carion2020detr}. This provides a set-level training signal, but inference remains parallel and relies on independent scoring.

\textbf{k-Means (oracle $k$).}
A non-learned clustering baseline that uses the ground-truth number of clusters and assigns each representative to the nearest centroid~\cite{lloyd1982kmeans}. While it has access to correct cardinality, it does not learn task-specific selection interactions.

\textbf{Autoregressive pointer network.}
A sequential model that selects representatives one at a time using a Transformer decoder and an EOS token~\cite{vinyals2015pointer,vaswani2017attention}. It models history-dependent selection but requires autoregressive decoding with step-wise recomputation.

\subsubsection{Results}
\label{sec:subset-results}

Table~\ref{tab:cluster_results} shows the central failure mode of parallel unary selection.
BCE recovers nearly all clusters, but does so by selecting many redundant elements, leading to low precision and large cardinality error.
Hungarian set prediction improves over BCE by using feature-space matching to choose plausible representatives, but its thresholded unary decoding still struggles to control coverage and cardinality.


\begin{table}[htbp]
    \centering
    \caption{Unordered structured selection on the representative subset-selection benchmark.
    Metrics are computed at the latent-cluster level.
    Runtime is reported as mean $\pm$ standard deviation.}
    \label{tab:cluster_results}
    \small
    \begin{tabular}{l c c c c c}
        \toprule
        \textbf{Method} & CluRec $\uparrow$ & CluPrec $\uparrow$ & CluF1 $\uparrow$ & CardErr $\downarrow$ & Runtime (ms) $\downarrow$ \\
        \midrule
        k-Means (oracle $K$) & 0.778 & 0.778 & 0.778 & 0.00 & -- \\
        BCE thresholding & \textbf{0.992} & 0.099 & 0.176 & 65.69 & 1.50 $\pm$ 0.11 \\
        Hungarian set prediction & 0.665 & 0.592 & 0.617 & 1.32 & \textbf{1.30 $\pm$ 0.08} \\
        Autoregressive pointer & 0.850 & \textbf{0.910} & \textbf{0.875} & \textbf{0.66} & 15.46 $\pm$ 0.22 \\
        \midrule
        \textbf{CPL} & 0.839 & 0.879 & 0.853 & 0.80 & 1.71 $\pm$ 0.07 \\
        \bottomrule
    \end{tabular}
\end{table}

Sequential conditioning resolves this tension.
The autoregressive pointer obtains the strongest final cluster-level scores, confirming that previous selections provide crucial information for avoiding duplicate clusters.
CPL attains similar cluster-level performance to AR while replacing full autoregressive decoding with a lightweight contextual update.
Oracle-$k$ k-Means provides a useful reference point: even with the correct subset size, fixed feature-space clustering underperforms learned history-dependent selection.

The runtime column highlights the same accuracy--runtime tradeoff observed in the ordered setting: CPL remains close to the thresholded parallel baselines and is much faster than full autoregressive decoding, while still retaining history-dependent selection.
The convergence curve in Appendix~\ref{app:training-convergence} further shows that this lightweight update also improves training efficiency: CPL reaches high cluster-level F1 earlier, even though AR eventually achieves the best final score.

\section{Related work}
\label{sec:related-work}

Our work builds on the Plackett--Luce family of discrete choice models, which represent distributions over rankings through sequential normalized selections~\citep{luce1959choice,plackett1975analysis}.
Classical formulations assume item-specific utilities that remain fixed during ranking construction~\citep{luce1959choice,plackett1975analysis,maystre2015plackettluce}.
Differentiable ranking methods relax discrete sorting or permutation operators to enable end-to-end optimization~\citep{grover2019stochastic,blondel2020fastsort}.

Beyond ranking, diversity-aware subset selection has been studied using models such as determinantal point processes, which encode repulsion through kernel determinants and have been widely used for diverse subset selection~\citep{kulesza2012dpp}.
These approaches define global probabilistic structures over sets but do not explicitly model history-dependent construction processes.

Permutation-invariant and permutation-equivariant architectures, including Deep Sets~\citep{zaheer2017deepsets} and Set Transformer~\citep{lee2019settransformer}, provide frameworks for encoding unordered inputs.
These models operate on sets but do not prescribe a decoding mechanism for constructing subsets under ambiguous supervision.

CPL is closely related to CRF-style models with unary and pairwise interactions, and more broadly to energy-based structured prediction~\citep{lafferty2001crf,krahenbuhl2011densecrf,belanger2016spen}.
However, such models often require dynamic programming or iterative approximate inference, such as message passing, sampling, mean-field updates, or continuous relaxation, when prediction depends on rich global interactions~\citep{krahenbuhl2011densecrf,belanger2016spen}.
By contrast, CPL adopts a Plackett--Luce factorization that enables efficient sequential decoding with precomputed interactions, avoiding repeated recomputation of deep representations.

Matching-based set prediction methods, most prominently DETR, formulate supervision as a bipartite assignment between predicted and target elements~\citep{kuhn1955hungarian,carion2020detr}.
Differentiable relaxations such as Gumbel--Sinkhorn networks extend this idea by approximating permutation and matching operations in a differentiable manner~\citep{mena2018gumbelsinkhorn}.
These approaches typically rely on parallel prediction followed by assignment or selection.

Autoregressive models construct outputs sequentially, where each decision conditions on previously generated elements.
Pointer Networks introduced this paradigm for selecting positions from an input sequence~\citep{vinyals2015pointer}, and subsequent neural combinatorial optimization methods adopt similar sequential policies~\citep{bello2016neural}.
Pix2Seq applies a related autoregressive formulation to object detection by representing visual outputs as token sequences~\citep{chen2022pix2seq}.
While expressive, these methods incur inference costs that scale with output length due to sequential dependence.

Finally, diffusion-based models offer a powerful framework for modeling multimodal distributions and generating diverse structured outputs~\citep{ho2020ddpm,janner2022diffuser,jiang2023motiondiffuser}.
However, they require iterative denoising with many network evaluations per sample, leading to high inference cost, and even accelerated variants remain iterative rather than single-pass~\citep{ho2020ddpm,song2021ddim}.
In contrast, CPL targets the setting where a single coherent structured output must be selected under ambiguity.
By combining parallel computation with lightweight sequential updates, CPL produces such outputs in a single forward pass followed by inexpensive decoding steps.
Additionally, CPL operates directly on discrete candidate sets and enforces structural constraints, such as non-repetition and termination, during decoding.

\label{sec:limitations}
\section{Limitations and future work}
\label{sec:limitations}

CPL is designed for structured selection problems where outputs can be constructed as sequences of discrete choices and dependencies are captured through unary and pairwise interactions. While effective for the settings studied in this work, this may be limiting for tasks requiring richer higher-order structure.

CPL produces a single output under greedy decoding, which is suitable when the goal is to return one coherent solution, but does not explicitly represent uncertainty or multiple diverse hypotheses. However, the underlying Plackett--Luce factorization naturally defines a conditional distribution over sequences, enabling efficient sampling-based decoding without recomputing representations.

Finally, CPL relies on a dense interaction matrix over candidates, which is practical at the scales considered here but may become challenging when the candidate set is very large. More scalable variants could use sparse or locality-aware interaction structures.

Future work includes extending CPL to higher-order interactions, scalable sparse parameterizations, and explicit multi-hypothesis decoding.

\section{Conclusion}
CPL provides a middle ground between parallel set prediction and fully autoregressive decoding by combining parallel construction of interaction parameters with lightweight history-dependent selection. Across both ordered and unordered settings, the results suggest that explicit sequential commitment is important for resolving ambiguity under incomplete supervision, while full autoregressive recomputation is not always necessary.

\newpage
\bibliographystyle{plain} 
\bibliography{references} 

\appendix
\newpage
\section{Visualizations of CPL selection process}
\label{appendix:greedy_decoding}

\begin{figure}[htbp]
    \centering
    \subcaptionbox{CPL Sequential Selection Steps\label{fig:cpl_steps_1}}{%
        \includegraphics[width=0.9\textwidth]{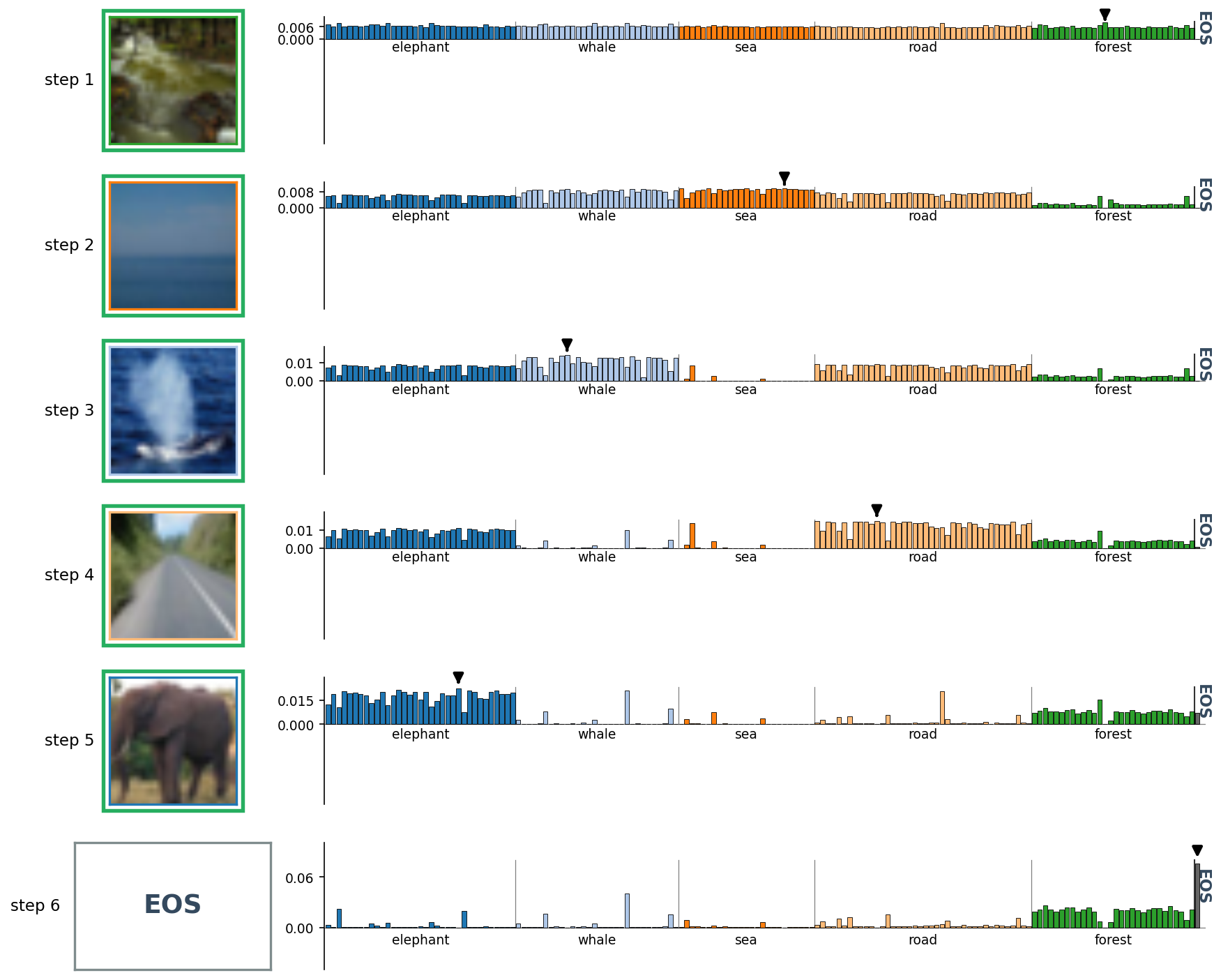}
    }

    \vspace{1em}

    \begin{minipage}{0.49\textwidth}
        \centering
        \subcaptionbox{Embedding Space\label{fig:tsne_1}}{%
            \includegraphics[width=\textwidth]{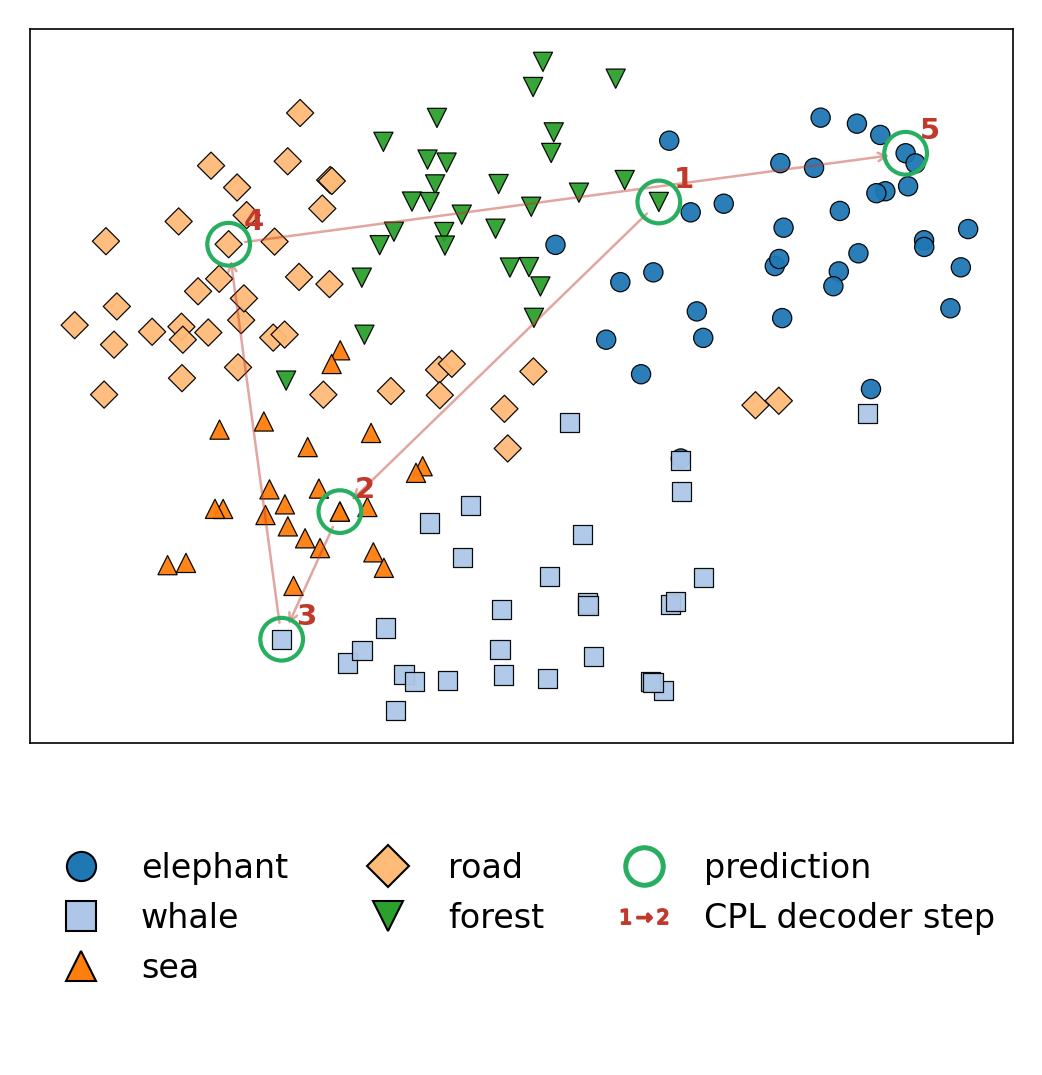}
        }
    \end{minipage}
    \hfill
    \begin{minipage}{0.49\textwidth}
        \centering
        \subcaptionbox{Matrix $W$\label{fig:w_matrix_1}}{%
            \includegraphics[width=\textwidth]{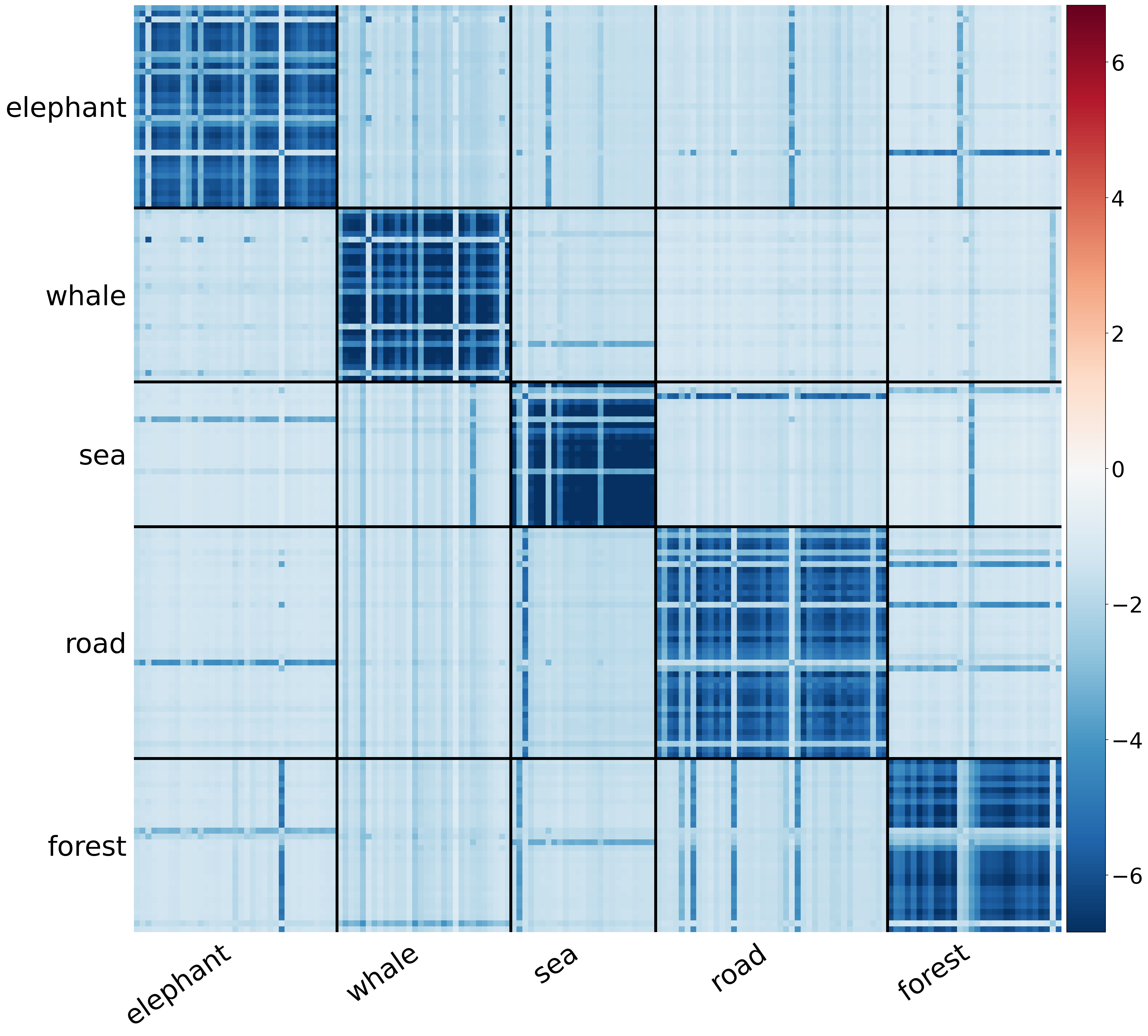}%
        }
    \end{minipage}

    \caption{\textbf{Example 1:} A selection example consisting of five semantic clusters: elephant, whale, sea, road, forest.}
    \label{fig:cpl_viz_1}
\end{figure}

\begin{figure}[htbp]
    \centering
    \subcaptionbox{CPL Sequential Selection Steps\label{fig:cpl_steps_2}}{%
        \includegraphics[width=0.92\textwidth]{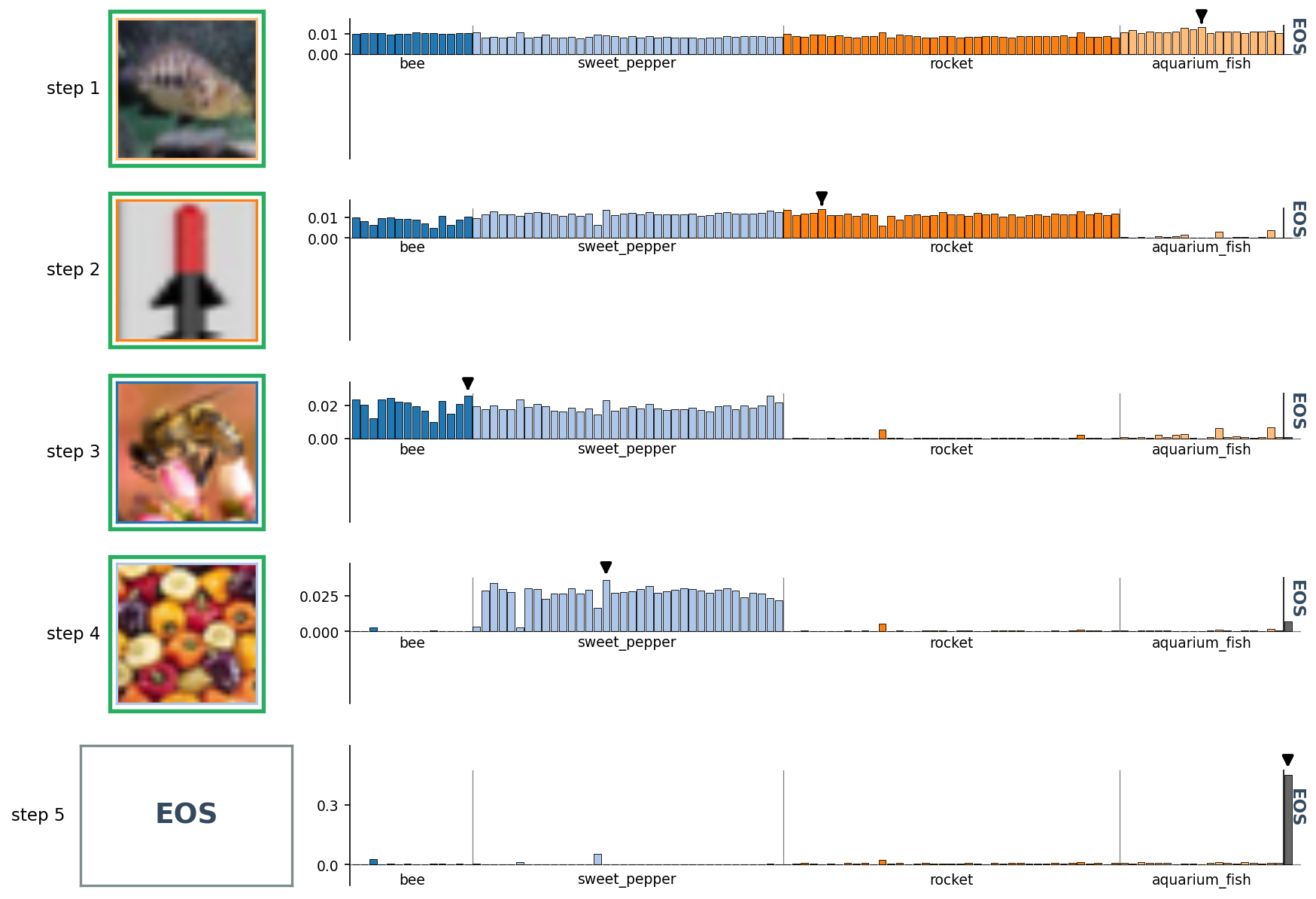}%
    }

    \vspace{1em}

    \begin{minipage}{0.49\textwidth}
        \centering
        \subcaptionbox{Embedding Space (Variant)\label{fig:tsne_alt_2}}{%
            \includegraphics[width=\textwidth]{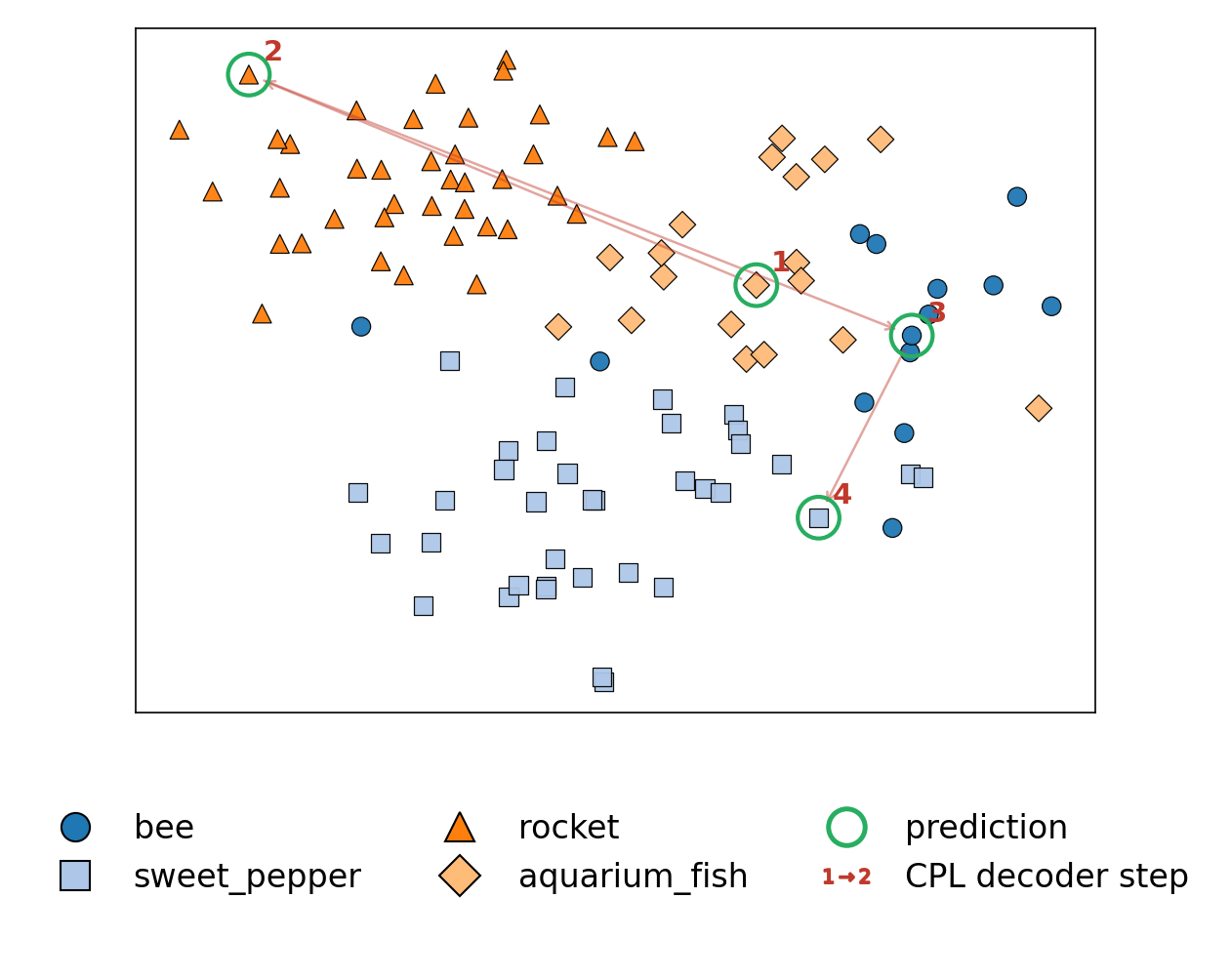}%
        }
    \end{minipage}
    \hfill
    \begin{minipage}{0.49\textwidth}
        \centering
        \subcaptionbox{Matrix $W$\label{fig:w_matrix_2}}{%
            \includegraphics[width=\textwidth]{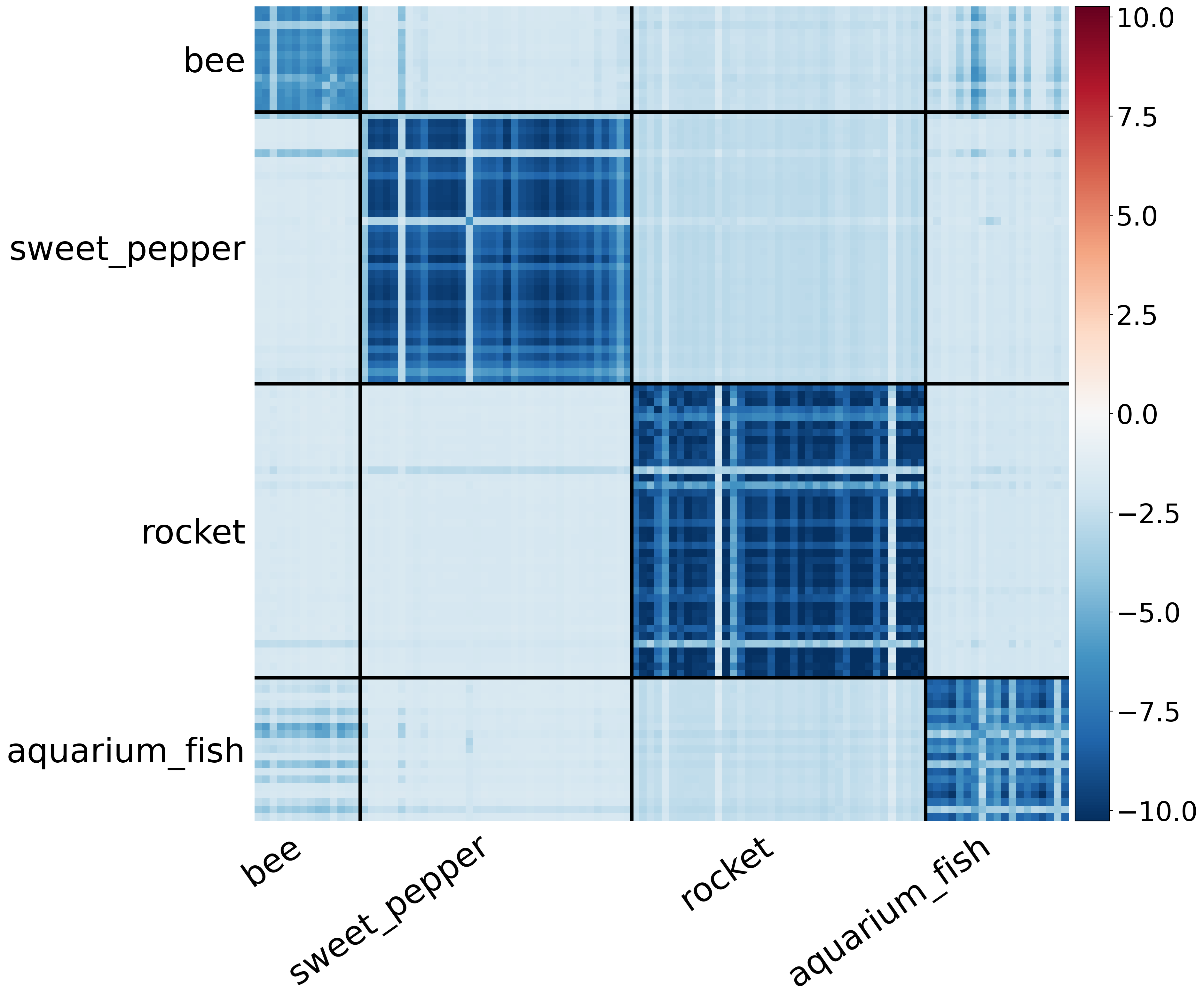}%
        }
    \end{minipage}

    \caption{\textbf{Example 2:} A selection example consisting of four semantic clusters: bee, sweet pepper, rocket, aquarium fish.}
    \label{fig:cpl_viz_2}
\end{figure}

In this section, we provide additional qualitative examples of the CPL selection process. Figures~\ref{fig:cpl_viz_1} and~\ref{fig:cpl_viz_2} illustrate how learned pairwise interactions induce structured, diversity-aware selection by sequentially choosing representative elements from different semantic clusters.

\paragraph{(a) Sequential selection dynamics.}
This panel visualizes the greedy decoding process of CPL. The selection begins by choosing the highest-probability element under the current model state. After each selection, the interaction structure suppresses the probabilities of elements within the same semantic cluster, reducing redundancy and encouraging exploration of alternative clusters. As a result, subsequent selections progressively shift across different clusters, producing a diverse set of representatives. The process terminates when the EOS token attains higher probability than any remaining candidate element, indicating that sufficient coverage has been achieved.

\paragraph{(b) Embedding space structure.}
This panel shows the frozen ResNet-18 feature embedding space. While the clusters are not perfectly separated in the raw representation space, CPL is still able to identify and select a single representative from each underlying semantic cluster. The numbers overlaid on the points indicate the order of selection during decoding, highlighting how the model traverses different regions of the embedding space over time.

\paragraph{(c) Interaction matrix $W$.}
This panel visualizes the learned interaction matrix $W$, which encodes pairwise dependencies between elements. The matrix exhibits a block-structured pattern, where each block corresponds to a semantic cluster. Strong negative interactions within blocks indicate mutual suppression among elements of the same cluster, while near-zero interactions between blocks indicate minimal coupling across clusters. This structure enables CPL to avoid redundant selections and promotes diversity across distinct semantic modes.

\newpage
\section{Path prediction visualization}
\label{path_prediction_viz}

\begin{figure}[htbp]
\centering

\textbf{(a)}\hspace{0.5em}
\begin{minipage}{0.29\textwidth}
    \centering
    \includegraphics[width=\textwidth, trim={70 70 50 0}, clip]{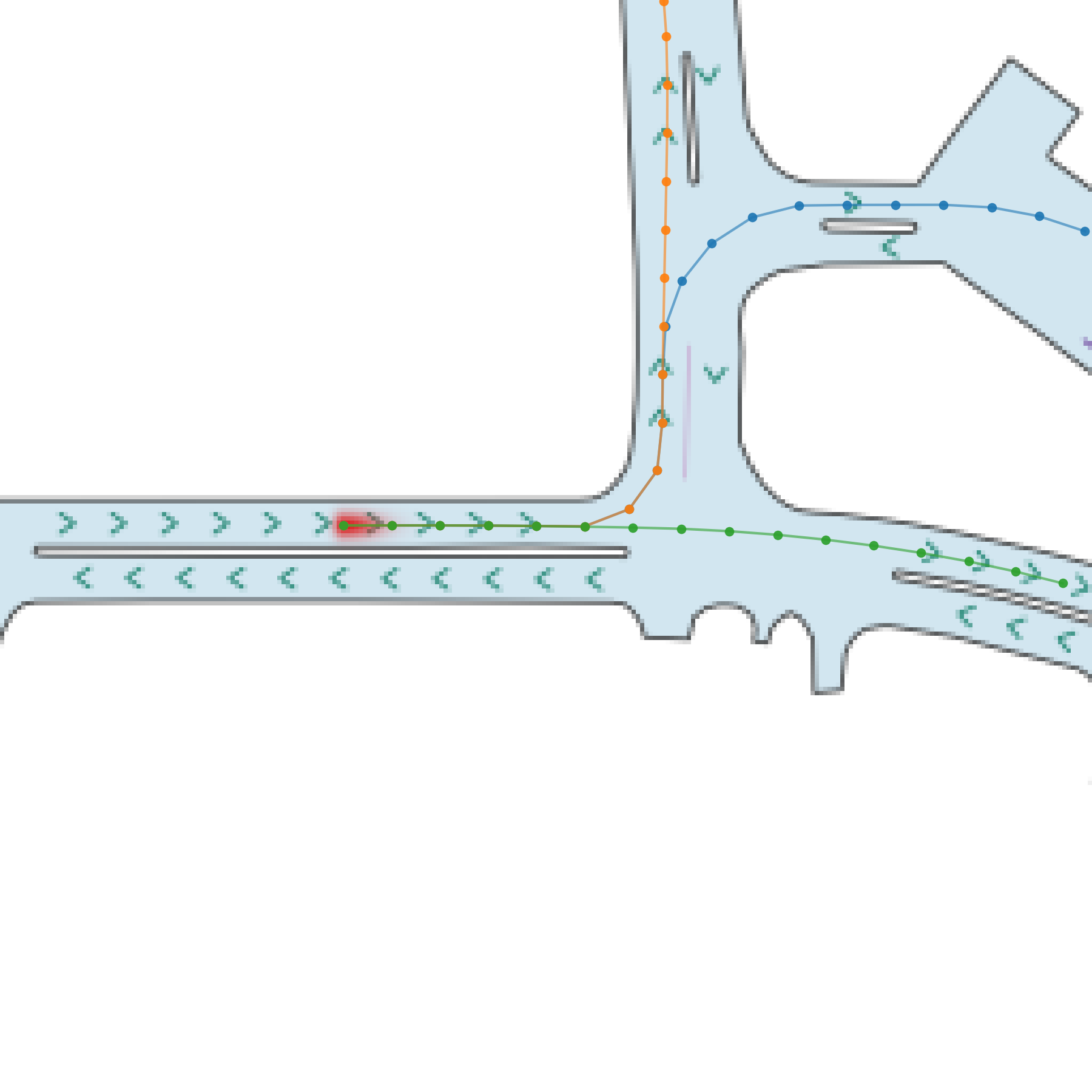}
\end{minipage}
\hfill
\begin{minipage}{0.29\textwidth}
    \centering
    \includegraphics[width=\textwidth, trim={70 70 50 0}, clip]{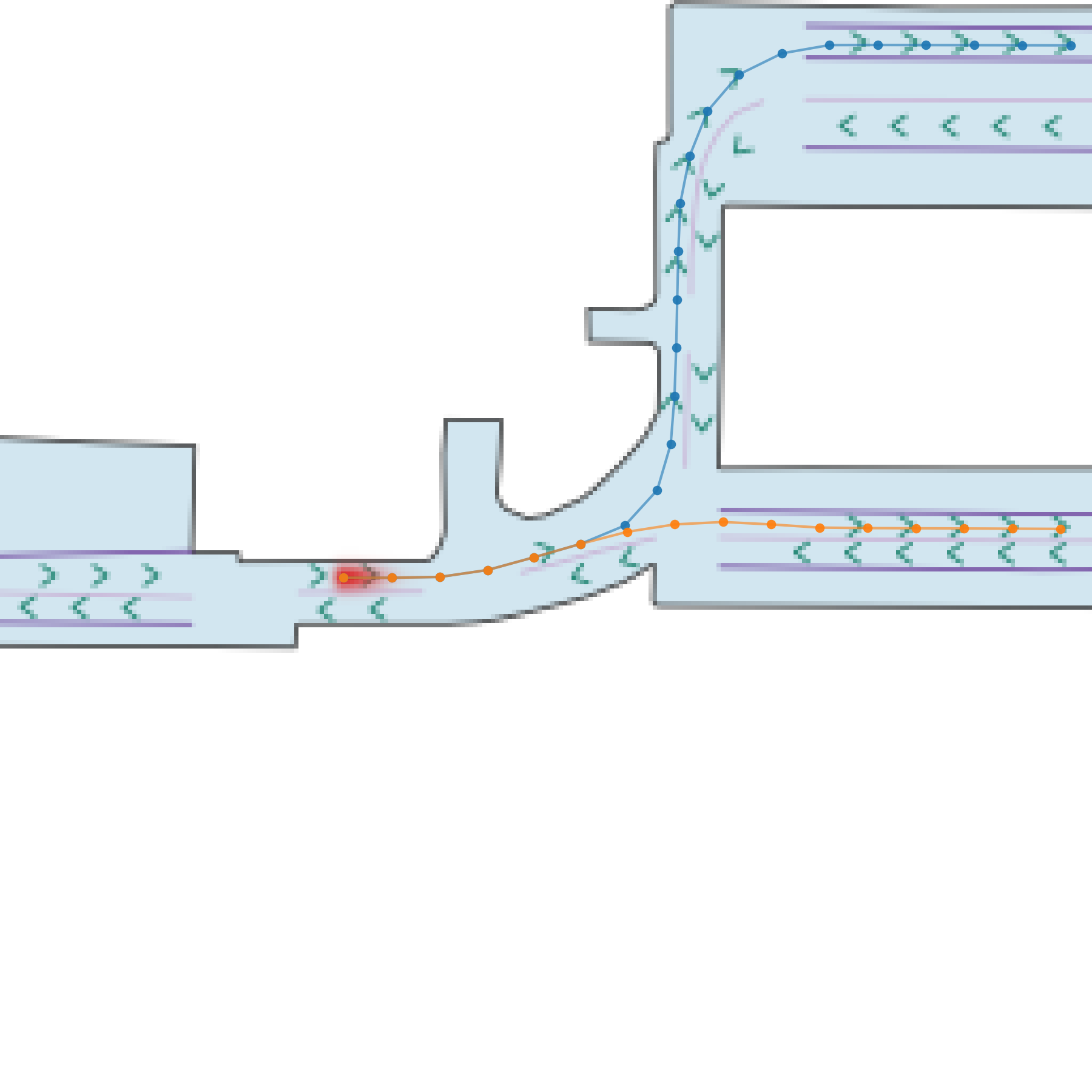}
\end{minipage}
\hfill
\begin{minipage}{0.29\textwidth}
    \centering
    \includegraphics[width=\textwidth, trim={70 70 50 0}, clip]{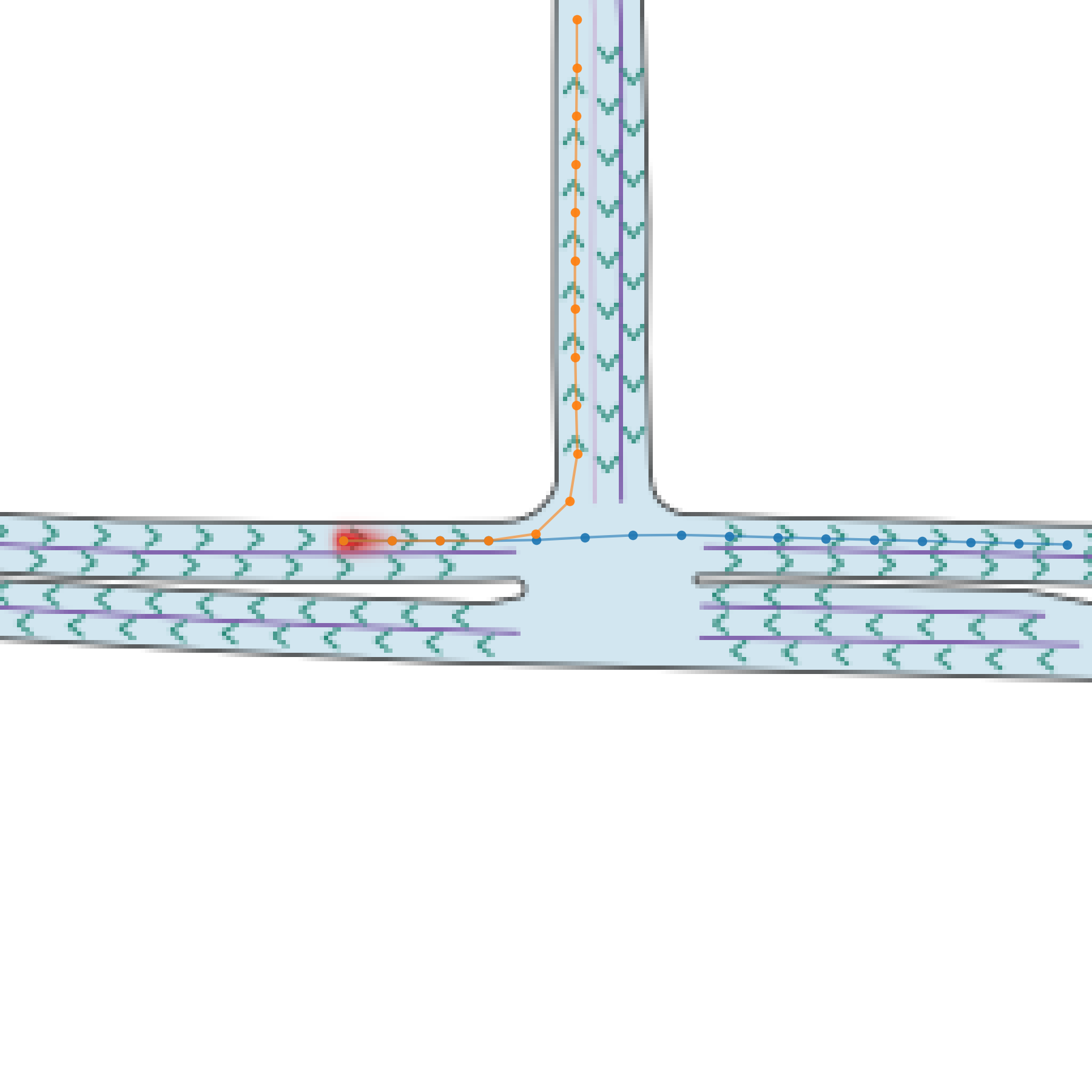}
\end{minipage}

\textbf{(b)}\hspace{0.5em}
\begin{minipage}{0.29\textwidth}
    \centering
    \includegraphics[width=\textwidth, trim={70 70 50 0}, clip]{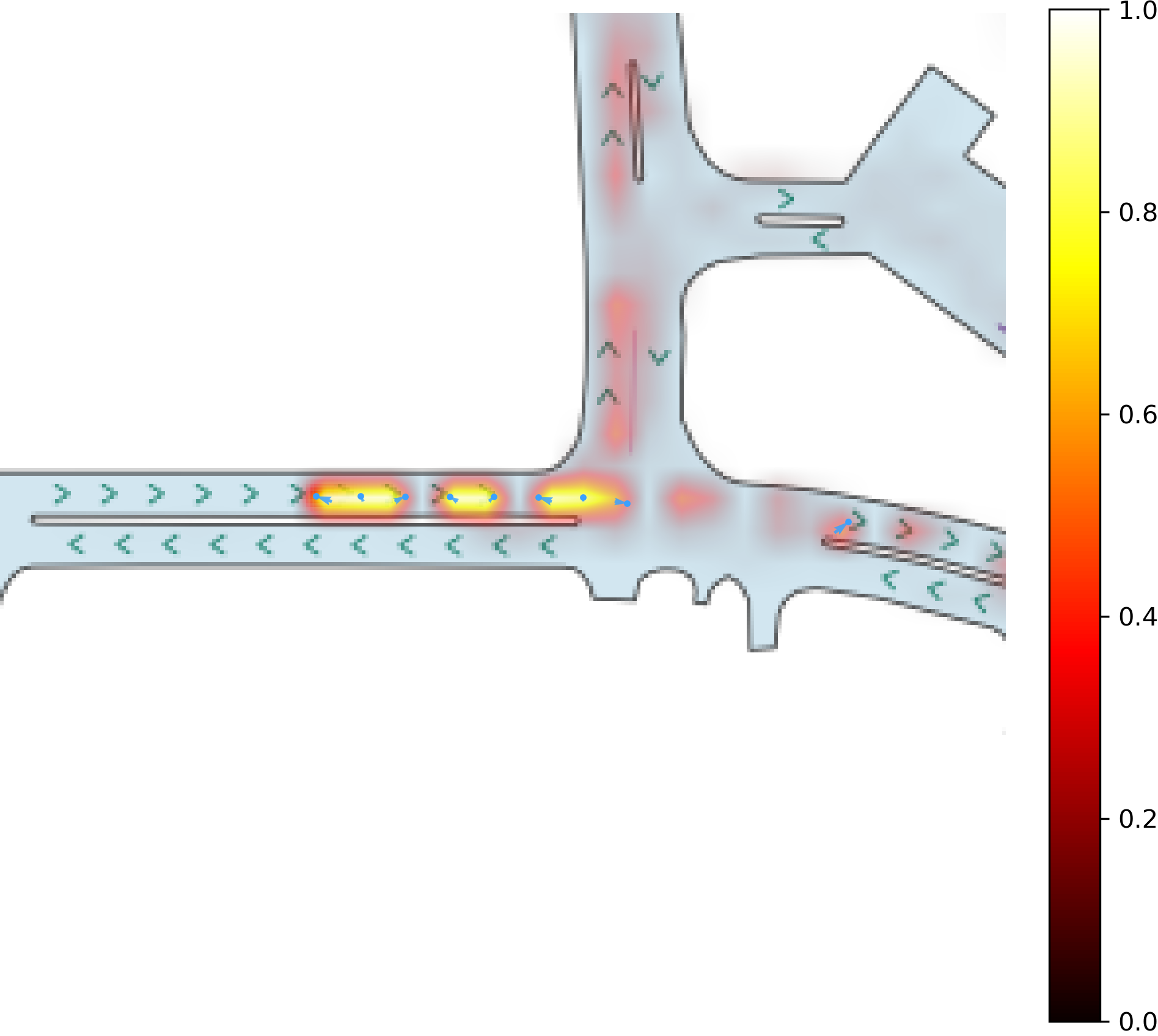}
\end{minipage}
\hfill
\begin{minipage}{0.29\textwidth}
    \centering
    \includegraphics[width=\textwidth, trim={70 70 50 0}, clip]{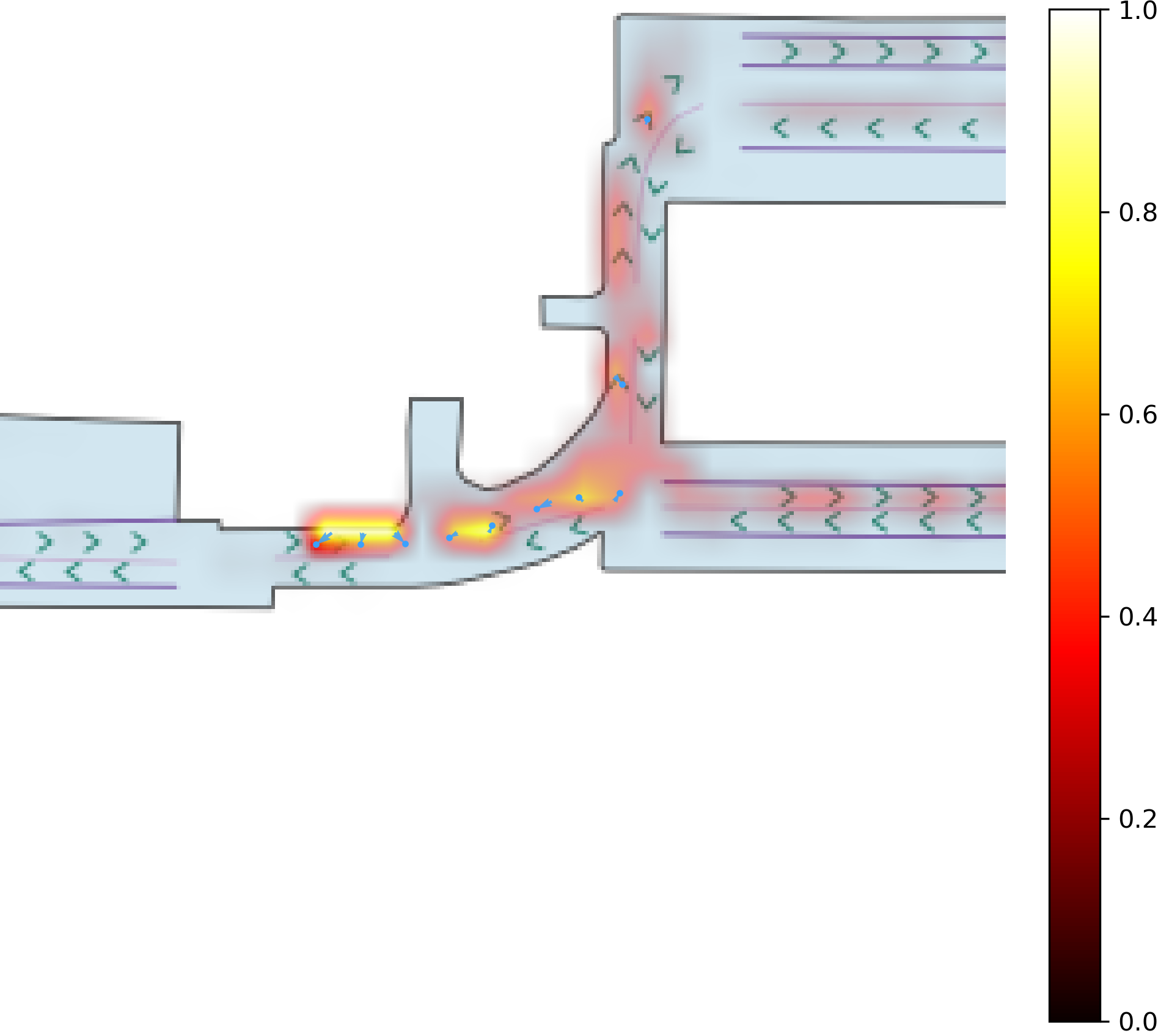}
\end{minipage}
\hfill
\begin{minipage}{0.29\textwidth}
    \centering
    \includegraphics[width=\textwidth, trim={70 70 50 0}, clip]{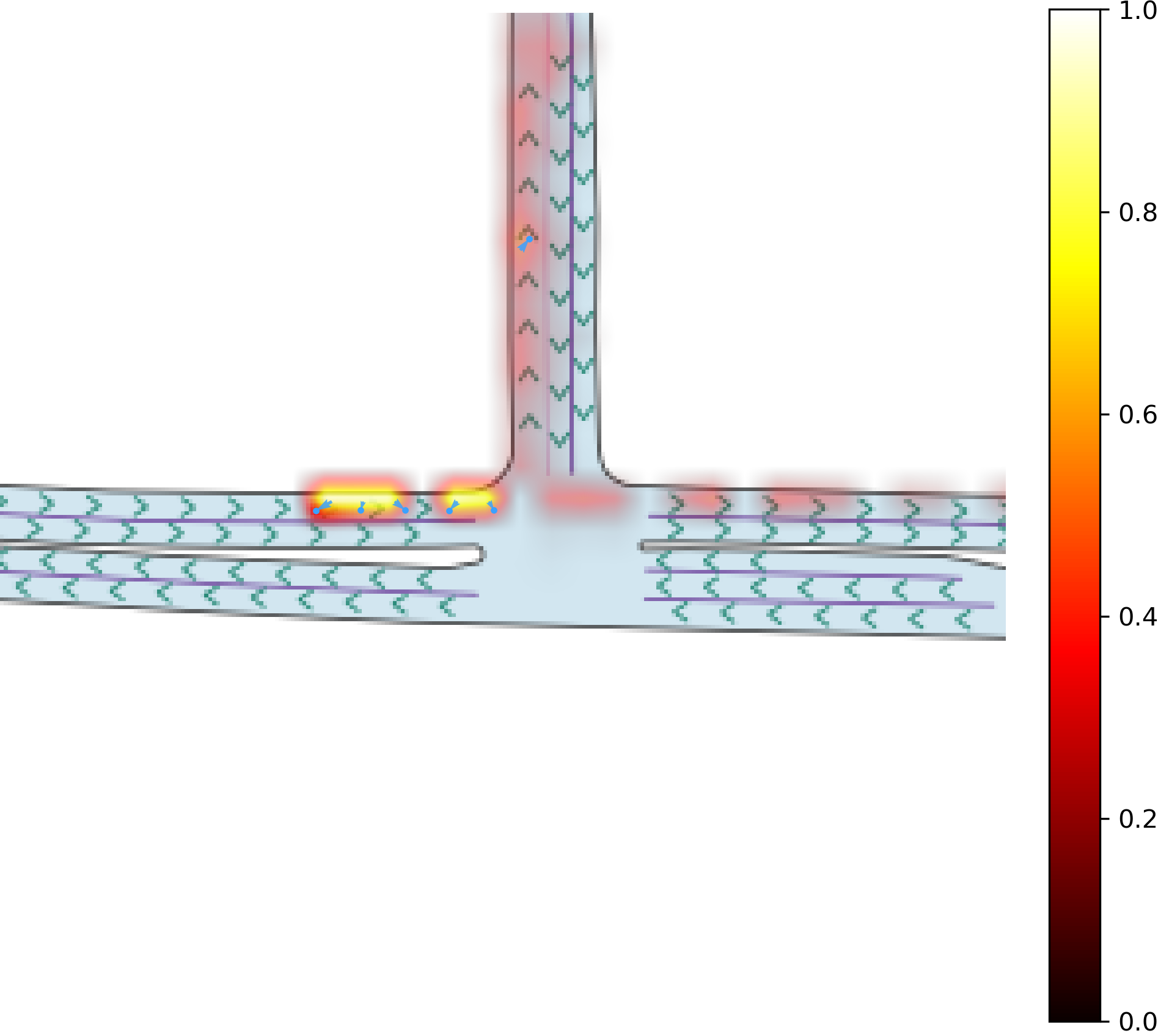}
\end{minipage}

\textbf{(c)}\hspace{0.5em}
\begin{minipage}{0.29\textwidth}
    \centering
    \includegraphics[width=\textwidth, trim={70 70 50 0}, clip]{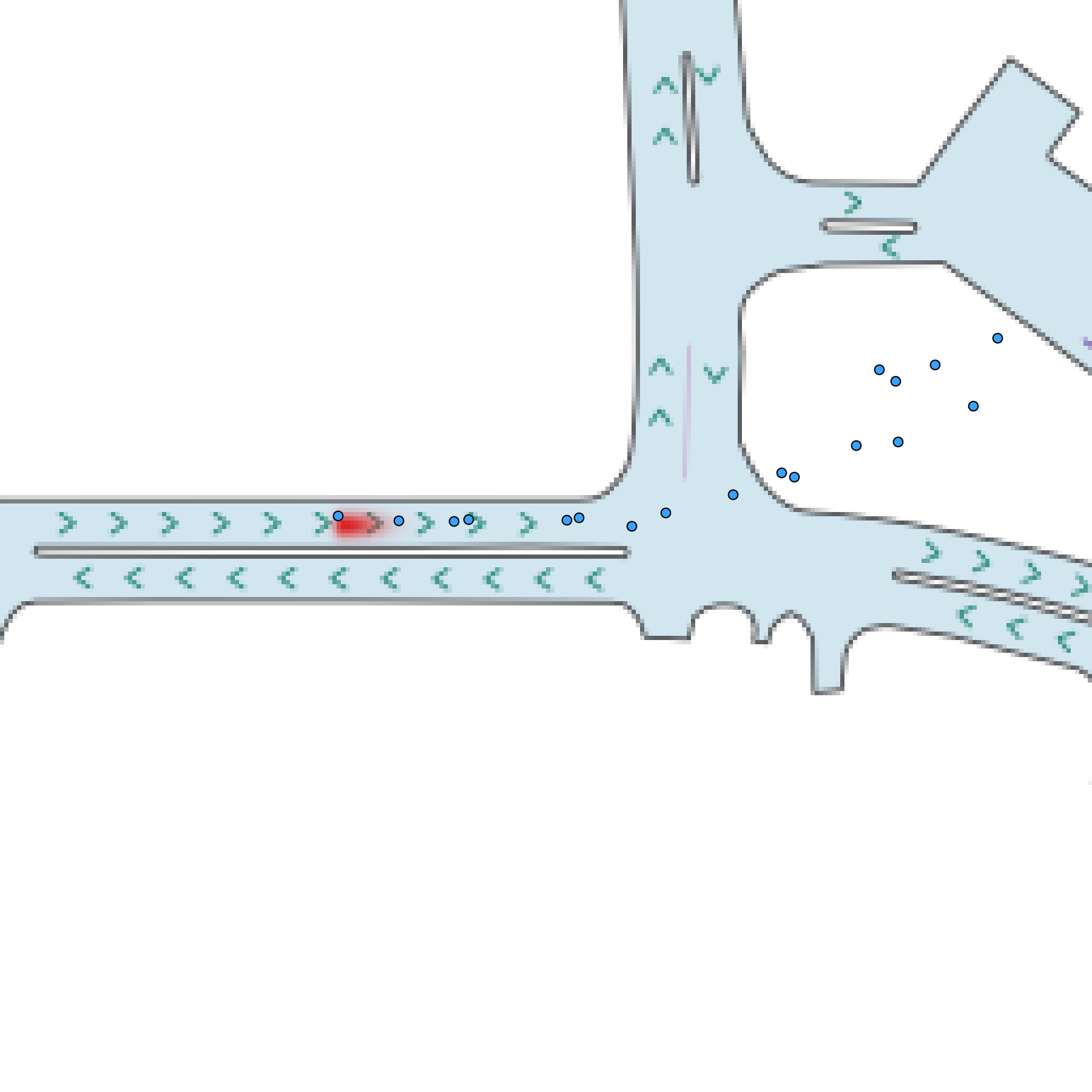}
\end{minipage}
\hfill
\begin{minipage}{0.29\textwidth}
    \centering
    \includegraphics[width=\textwidth, trim={70 70 50 0}, clip]{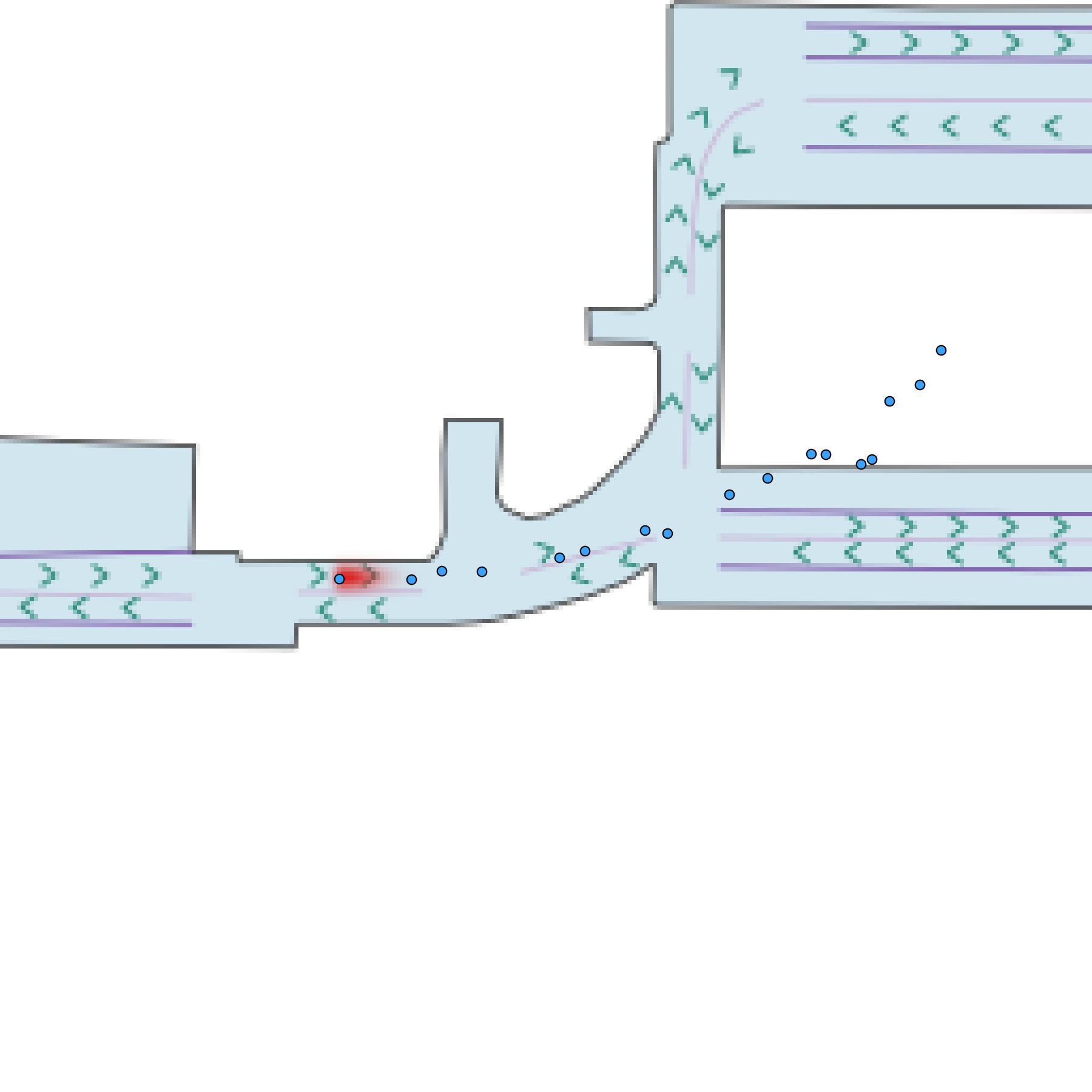}
\end{minipage}
\hfill
\begin{minipage}{0.29\textwidth}
    \centering
    \includegraphics[width=\textwidth, trim={70 70 50 0}, clip]{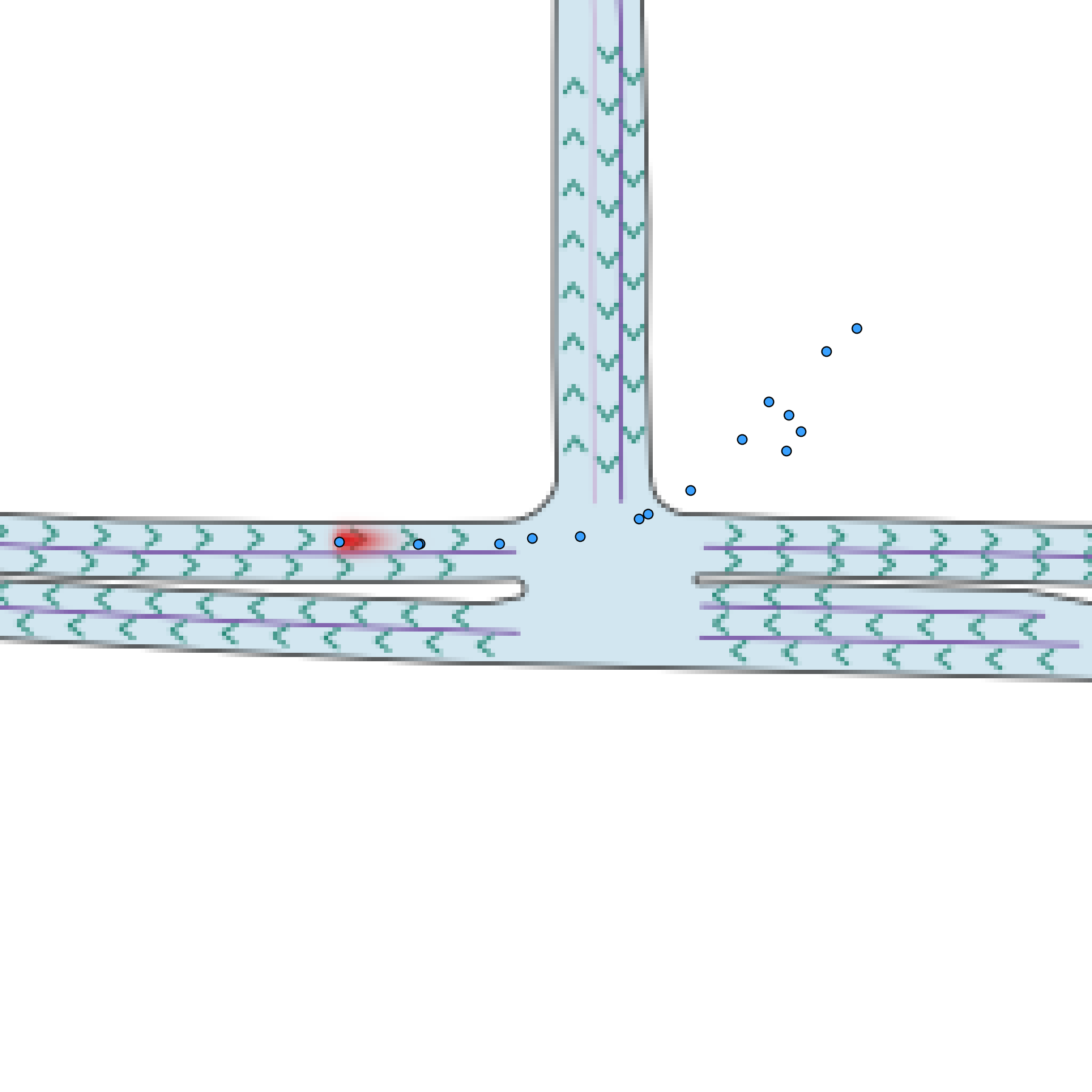}
\end{minipage}

\textbf{(d)}\hspace{0.5em}
\begin{minipage}{0.29\textwidth}
    \centering
    \includegraphics[width=\textwidth, trim={70 70 50 0}, clip]{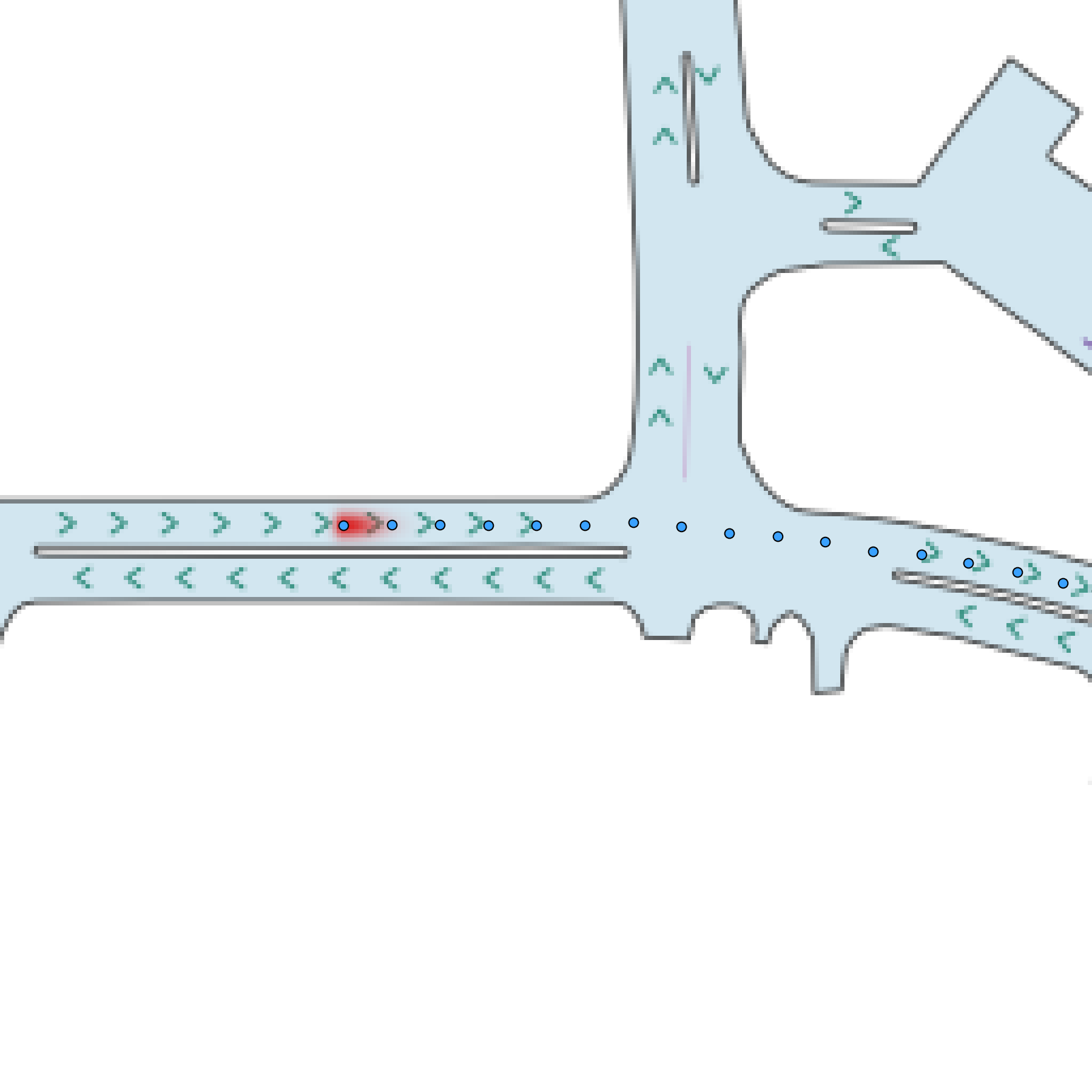}
\end{minipage}
\hfill
\begin{minipage}{0.29\textwidth}
    \centering
    \includegraphics[width=\textwidth, trim={70 70 50 0}, clip]{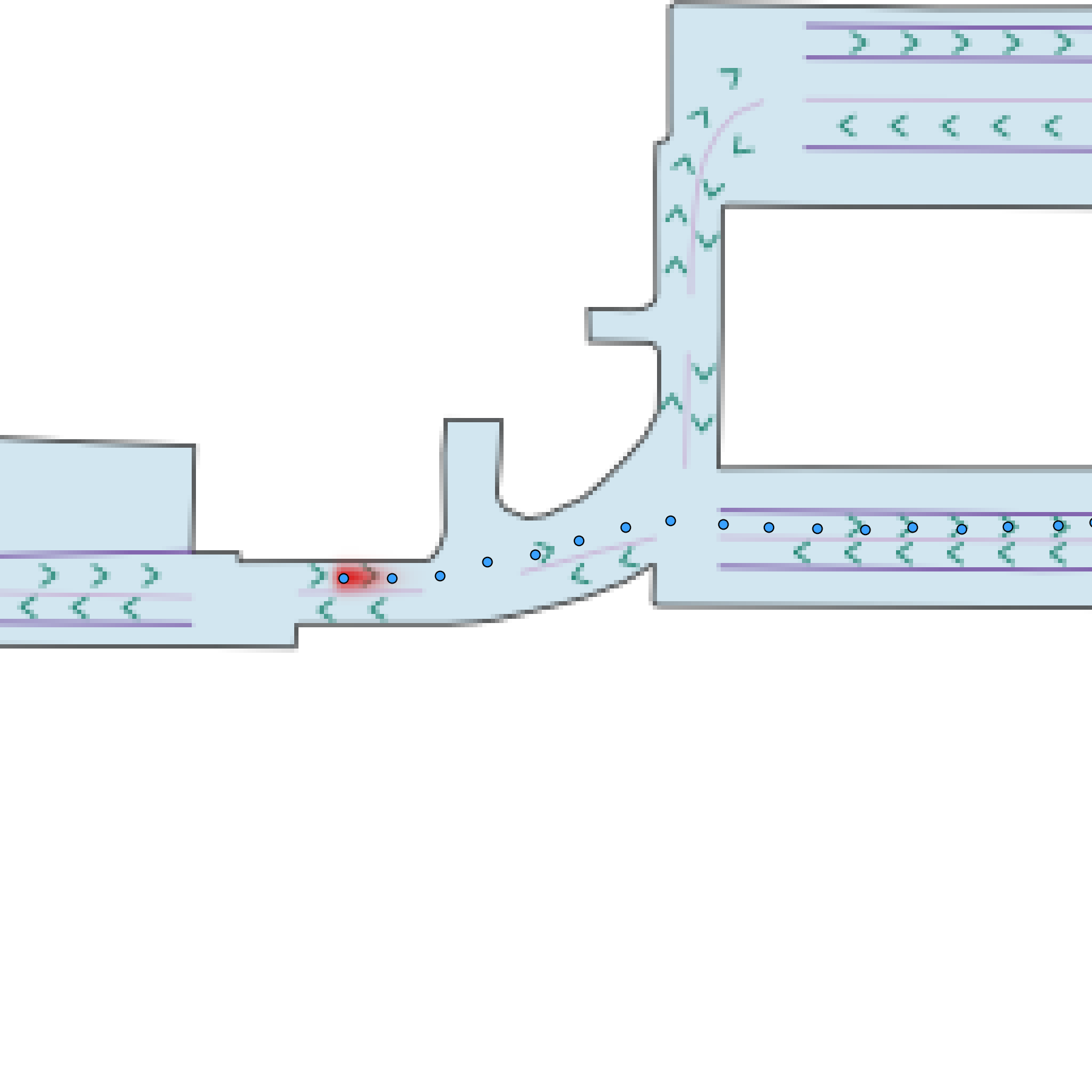}
\end{minipage}
\hfill
\begin{minipage}{0.29\textwidth}
    \centering
    \includegraphics[width=\textwidth, trim={70 70 50 0}, clip]{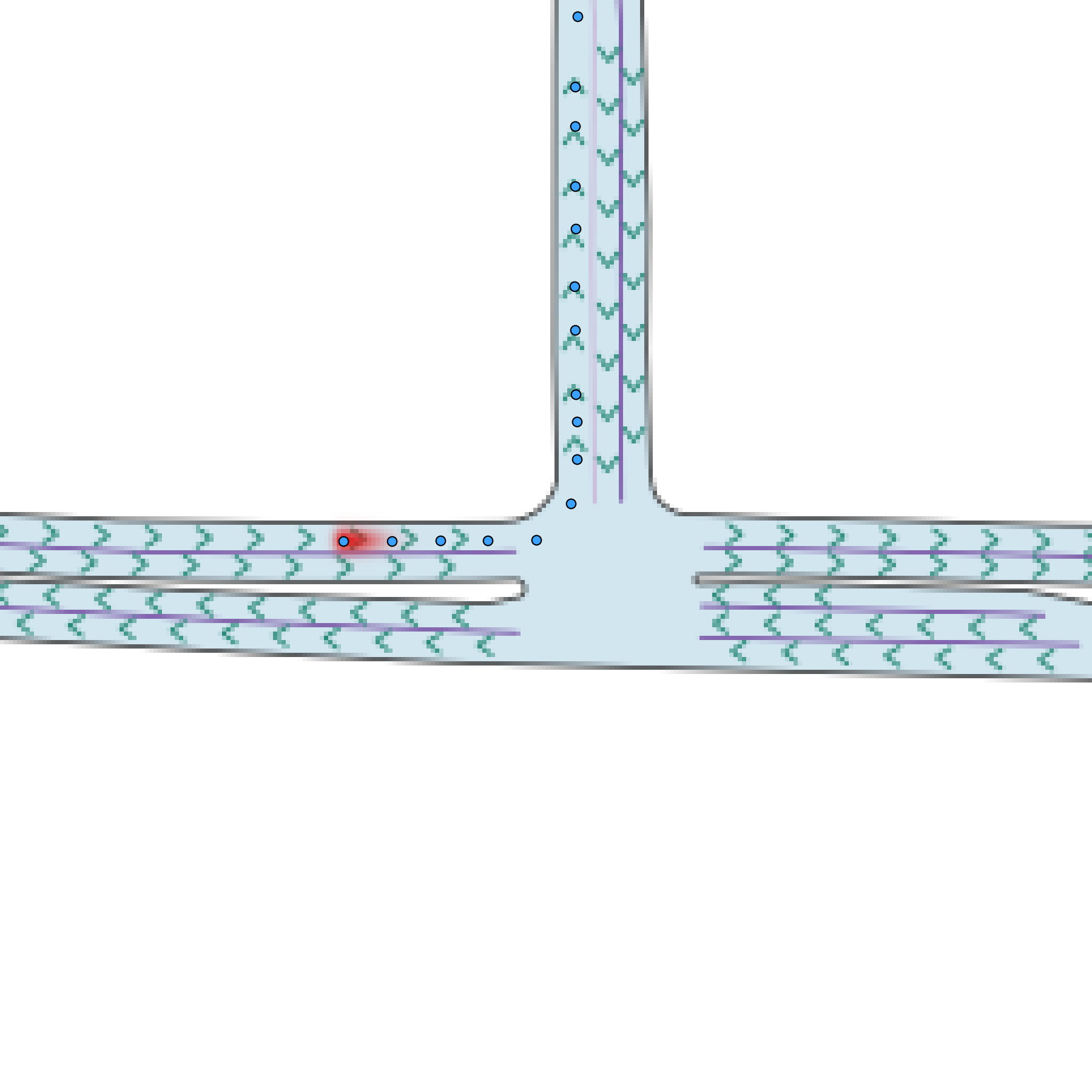}
\end{minipage}

\caption{Path prediction visualization. Each column corresponds to a different scene instance. (a) Set of all valid ground-truth driving trajectories. (b) Predictions obtained using an independence-based scoring method, which may lead to fragmented paths. (c) Predictions obtained via Hungarian matching, resulting in an averaged path. (d) Predictions generated using the CPL-based decoding strategy.}
\label{fig:path_prediction_grid}
\end{figure}

\newpage
\section{Implementation details}

\paragraph{CPL head implementation.}
Across both experiments, the CPL head receives encoded candidate embeddings and produces unary scores and a dense pairwise interaction matrix.
The unary function $f_\theta$ is implemented as an MLP applied independently to each candidate and to a learned EOS token.
The pairwise function $f_W$ is implemented using two separate MLP projections, producing key and value vectors for each candidate.
The interaction matrix is computed by a scaled dot product,
\[
W_{ji} = \frac{k_j^\top v_i}{\sqrt{d_h}},
\]
so column $i$ of $W$ gives the logit update applied after candidate $i$ is selected.
The EOS row and column of $W$ are set to zero, allowing EOS to terminate decoding without contributing pairwise context.

\subsection{Path prediction implementation details}
\label{app:path-prediction-details}

This appendix records the main settings needed to reproduce the multi-modal path prediction experiment.
We include only parameters that determine the benchmark, model capacity, training recipe, evaluation protocol, or differences between compared variants.

\paragraph{Dataset generation.}
The benchmark is generated from local road graphs extracted from real-world nuScenes maps~\citep{caesar2020nuscenes}, including drivable areas, lane dividers, and road boundaries.
The generated scenes cover forks, intersections, curved segments, and an ego pose.
Each sample covers a $128 \times 128$ meter region and is rasterized as a BEV RGB map at $0.5$ meter per pixel resolution.
Images have resolution $256 \times 256$ and are discretized with downsampling factor $D=8$, yielding a $32 \times 32$ decision grid.
The dataset contains $6{,}000$ samples with a deterministic $80/20$ train/validation split.
Each scene stores up to $K_{\max}=6$ valid paths, and each path is capped at $M_{\max}=64$ sparsified points after arc-length resampling and per-cell deduplication.

\paragraph{Shared architecture.}
All learned methods use the same shared encoder and offset prediction head.
The encoder consists of a ResNet-18 backbone~\citep{he2016resnet} followed by a Transformer encoder~\citep{vaswani2017attention} with hidden dimension $128$, $4$ layers, $8$ attention heads, dropout $0.1$, and GELU feed-forward blocks~\citep{hendrycks2016gelu}.
The heatmap and offset heads are per-cell MLPs with hidden width $64$.
The compared variants differ only in the selection mechanism: independent heatmap scoring, Hungarian matched heatmap scoring~\citep{kuhn1955hungarian,carion2020detr}, a parallel multi-hypothesis head~\citep{chai2020multipath,phanminh2020covernet}, an autoregressive pointer decoder~\citep{vinyals2015pointer}, or the CPL head.
The CPL head follows the shared implementation described above.

\paragraph{Training.}
Models are trained with AdamW~\citep{loshchilov2019adamw} using learning rate $3 \times 10^{-4}$, weight decay $10^{-4}$, batch size $256$, and $300$ epochs.
Horizontal flip augmentation is enabled.
The default grid-supervised loss uses heatmap BCE and an $\ell_1$ offset loss, with weights $10$ and $5$, respectively.
For CPL and AR under grid matching, the sequence loss replaces the heatmap classification term while the offset loss remains supervised on the selected grid cells.
For each variant, we select the checkpoint with the best validation $\min$-$\mathrm{ADE}$ and report final validation metrics for that checkpoint.

\paragraph{Evaluation and runtime.}
Metrics are computed in full-resolution pixel coordinates.
For each prediction, $\min$-$\mathrm{ADE}$ and $\min$-$\mathrm{HD}$ are reduced over all valid ground-truth paths in the scene, so a model is rewarded for matching any valid continuation~\citep{huttenlocher1993hausdorff}.
Off-road rate is measured by querying predicted points against the binarized road channel.
Runtime is reported as mean per-example inference time in milliseconds, including the shared encoder, heads, and any method-specific decoding loop.
All training runs and runtime measurements were performed on an NVIDIA A100 GPU.
Runtime was measured with batch size $1$.
Each training run took approximately 2.5 hours on a single NVIDIA A100 GPU and used about 30 GB of GPU memory.
\subsection{Representative subset selection implementation details}
\label{appendix:cifar_clustering_details}

This appendix records the main settings needed to reproduce the representative subset selection experiment.
We include only parameters that define the benchmark, model capacity, training recipe, evaluation protocol, or meaningful differences between compared variants.

\paragraph{Dataset construction.}
The reported experiment uses CIFAR-100~\citep{krizhevsky2009cifar}.
Images are resized to $224 \times 224$, normalized with ImageNet statistics~\citep{russakovsky2015imagenet}, and embedded with a frozen ImageNet-pretrained ResNet-18~\citep{he2016resnet} whose final classification layer is removed.
Training bags are sampled from the CIFAR training split and validation bags from the CIFAR test split.
The default benchmark contains 5{,}000 training bags and 1{,}000 validation bags.
Each bag samples between 3 and 10 CIFAR classes and 12 to 40 images per sampled class, with the final bag capped at 160 elements.
After this cap, only classes still present in the bag define the target clusters.

\paragraph{Ambiguous supervision.}
Each time a bag is drawn, the ground-truth subset is resampled by selecting one random token from every present class.
Thus, the same latent clustering can induce many valid target subsets across epochs.
Evaluation never requires matching these sampled token identities, and instead uses the true class membership only to measure cluster coverage, duplicate selections, and cardinality error.

\paragraph{Shared architecture and optimization.}
All learned methods consume the frozen ResNet features and use a Transformer encoder~\citep{vaswani2017attention} with hidden dimension 512, 3 layers, 4 attention heads, feed-forward width 1024, GELU activations~\citep{hendrycks2016gelu}, dropout 0.1, and pre-norm LayerNorm~\citep{ba2016layernorm}.
MLP heads use hidden width 256.
Models are trained for 300 epochs with Adam~\citep{kingma2014adam}, learning rate $5 \times 10^{-4}$, batch size 256, and validation after every epoch.
Checkpoints are selected by validation CluF1: for CPL and AR this is the greedy decoding row, while for BCE and Hungarian this is the best row in the threshold sweep.

\paragraph{Variant-specific settings.}
BCE trains the shared encoder and unary head with masked binary cross-entropy, with the positive weight set automatically from the per-batch negative-to-positive ratio.
Hungarian uses the same unary model, but constructs targets with bipartite matching between valid tokens and the sampled representatives~\citep{kuhn1955hungarian,carion2020detr}.
The reported setting uses classification cost weight 10, squared-$\ell_2$ distance on frozen ResNet features with distance weight 0.01, excludes identity matches when an alternative exists, and does not use the optional consistency or entropy terms.
CPL uses the shared CPL head described above, with MLP hidden width 256.

CPL and AR share the same EOS-weighted sequence objective: 10 random target permutations are averaged per batch, early EOS is penalized with weight 5, and the final EOS target uses weight 1.
Greedy decoding is capped at 20 selections.
The AR baseline uses the same encoder, a 2-layer Transformer decoder~\citep{vaswani2017attention} with 4 attention heads and feed-forward width 1024, and a pointer-network readout~\citep{vinyals2015pointer} over the bag plus EOS.

\paragraph{Evaluation and runtime.}
BCE and Hungarian are decoded by sweeping sigmoid thresholds $\{0.1,0.2,\ldots,0.9,0.95\}$, and the table reports the best validation CluF1 among these rules.
CPL and AR decode greedily until EOS or the maximum selection budget.
The k-Means reference runs Lloyd k-Means~\citep{lloyd1982kmeans} directly on the frozen bag features with oracle access to the number of present semantic clusters, then selects the member nearest to each centroid.
Metrics are averaged over validation batches and reported as CluRec, CluPrec, CluF1, and CardErr.
Runtime is reported as mean per-example inference time in milliseconds.
All runtime measurements were performed on an NVIDIA A100 GPU with batch size 1.
Each training run took approximately 3.5 hours on a single NVIDIA A100 GPU and used about 5.5 GB of GPU memory.

\newpage
\section{Training convergence}
\label{app:training-convergence}

\begin{figure}[h]
    \centering
    \includegraphics[width=0.58\linewidth]{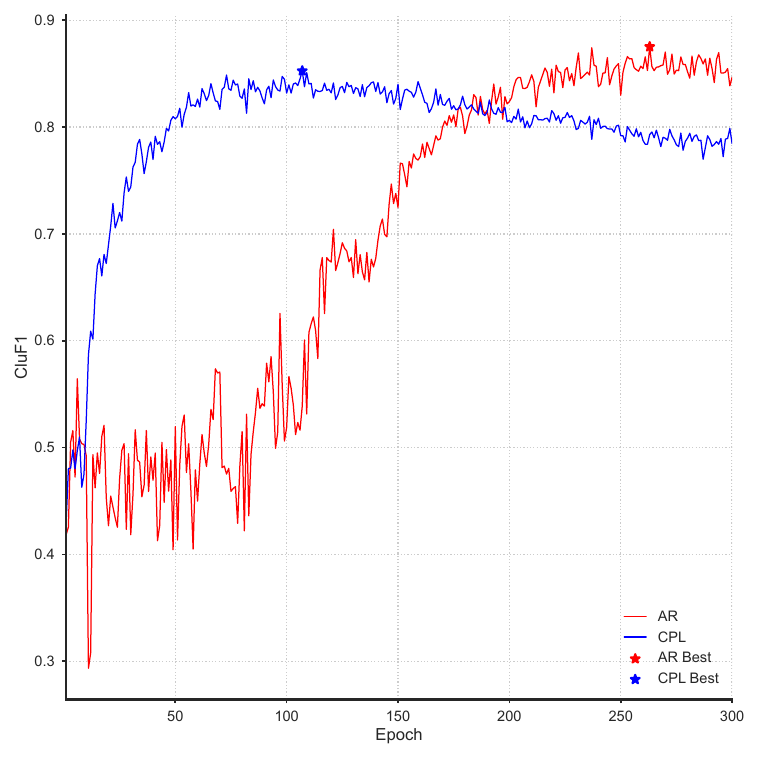}
    \caption{Training convergence on the representative subset selection task. Validation CluF1 is shown over training epochs for CPL and the autoregressive pointer network, with star marking the best validation score reached by each method.}
    \label{fig:convergence}
\end{figure}

Figure~\ref{fig:convergence} reports the validation CluF1 during training for CPL and the autoregressive pointer network on the representative subset selection task.

Although the autoregressive pointer network eventually achieves the strongest final score, CPL reaches competitive performance much earlier while retaining most of the benefit of history-dependent selection.
This is consistent with CPL's lighter parameterization, which provides history-dependent updates without requiring a full autoregressive decoder at every step.

\newpage
\section{Illustrative example of failure under ambiguous supervision}
\label{app:ambiguity_example}

We provide a minimal synthetic example illustrating how matching-based objectives can favor invalid averaged solutions under ambiguous supervision, and how CPL avoids this failure through sequential commitment.

\paragraph{Setup.}
Consider a candidate set of five elements:
\[
\{a, b, c, d, e\}.
\]
The ground-truth subset is ambiguous:
\[
S^\star =
\begin{cases}
\{a\}, & \text{with probability } \tfrac{1}{2}, \\
\{b, c\}, & \text{with probability } \tfrac{1}{2}.
\end{cases}
\]
Thus, there are two valid modes: a single-element representation $\{a\}$ and a two-element decomposition $\{b,c\}$.
Elements $d,e$ are never part of the ground truth, but can be interpreted as intermediate or hybrid candidates.

\paragraph{Matching-based objective.}
We consider a standard matching-based loss, where a predicted set $\hat{S}$ is matched to the ground-truth set using a cost function $\Delta(\cdot,\cdot)$, and unmatched elements incur a penalty of $1$.
We define pairwise matching costs:
\[
\Delta(a,b) = \Delta(a,c) = 1,\quad
\Delta(a,d) = \Delta(a,e) = \tfrac{1}{4},\quad
\Delta(b,d) = \Delta(c,e) = \tfrac{1}{3},
\]
with all other mismatches assigned cost $1$.

\paragraph{Interpretation of the cost.}
This setup can be interpreted as a simplified object detection scenario with ambiguous annotations. For example, an object composed of two parts may sometimes be labeled as a single entity and sometimes as two separate objects. Selecting $\{a\}$ corresponds to the single-object interpretation, while selecting $\{b,c\}$ corresponds to the two-part decomposition. The elements $d$ and $e$ represent intermediate or partial detections: $d$ captures $b$ together with part of $c$, and $e$ captures $c$ together with part of $b$. The matching costs reflect this structure, assigning lower cost to partial matches than to completely unrelated elements.

\paragraph{Failure under ambiguity.}
We compare three candidate predictions.

\emph{Predicting $\{a\}$.}
The cost is $0$ if $S^\star = \{a\}$ and $2$ if $S^\star = \{b,c\}$, yielding expected cost:
\[
\mathbb{E}[\Delta] = \tfrac{1}{2}(0) + \tfrac{1}{2}(2) = 1.
\]

\emph{Predicting $\{b,c\}$.}
By symmetry, the expected cost is also $1$.

\emph{Predicting $\{d,e\}$.}
If $S^\star = \{a\}$:
\begin{itemize}
    \item match $d$ to $a$: cost $\tfrac{1}{4}$,
    \item $e$ is unmatched: cost $1$,
\end{itemize}
total $1.25$.
If $S^\star = \{b,c\}$:
\begin{itemize}
    \item match $d$ to $b$: cost $\tfrac{1}{3}$,
    \item match $e$ to $c$: cost $\tfrac{1}{3}$,
\end{itemize}
total $\tfrac{2}{3}$.
Thus, the expected cost is:
\[
\mathbb{E}[\Delta] = \tfrac{1}{2}(1.25) + \tfrac{1}{2}\left(\tfrac{2}{3}\right)
= \tfrac{23}{24} \approx 0.96.
\]

\paragraph{Interpretation.}
Although $\{d,e\}$ is not a valid solution under any mode, it achieves lower expected loss than either valid subset.
This occurs because a deterministic matching-based predictor trained under expected set cost can favor intermediate configurations that partially match multiple modes, leading to structurally invalid outputs.

\paragraph{Why CPL avoids this failure.}
CPL models subset construction as a sequence of conditional decisions:
\[
P(S) = \prod_t P(i_t \mid S_{<t}),
\]
with logits:
\[
\ell_j(S_t) = \theta_j + \sum_{i \in S_t} W_{ji}.
\]
By committing to a choice early, CPL conditions future selections on previous ones, enabling consistent mode selection.

Importantly, pairwise interactions are sufficient to represent this behavior in the toy setting above.
In this example, one can construct parameters $(\theta, W)$ such that CPL produces only the valid subsets $\{a\}$ and $\{b,c\}$ under greedy decoding.
Intuitively, selecting $a$ suppresses the logits of $b,c$, while selecting $b$ promotes selecting $c$ and suppresses $a$, leading to coherent mode commitment.
A concrete construction is provided below.

\paragraph{Example construction.}
One possible parameterization is:
\[
\theta_a = \theta_b = 0.5,\quad \theta_c = \theta_d = \theta_e = 0,\quad \theta_{\mathrm{EOS}} = 0.1.
\]

We define the nonzero pairwise interactions as follows, where \(W_{ji}\) denotes the effect of previously selected element \(i\) on the logit of candidate \(j\):
\[
W_{j,a} = -1 \quad \forall j \in \{b,c,d,e\},
\]
\[
W_{c,b} = +1,\qquad W_{a,b} = -1,\qquad W_{a,c} = -1.
\]
All other interactions are set to zero, including all interactions into the EOS token.

Under this construction:
\begin{itemize}
    \item If $a$ is selected first, all remaining logits drop below EOS, yielding $\{a\}$.
    \item If $b$ is selected first, the interaction $W_{c,b}=+1$ makes $c$ the next selection, after which EOS is selected, yielding $\{b,c\}$.
\end{itemize}
Thus, greedy decoding produces only the valid subsets $\{a\}$ and $\{b,c\}$, and never selects $\{d,e\}$.

\paragraph{Summary.}
This example highlights a difference between the two formulations in this toy setting: a deterministic matching-based predictor optimized under expected loss can prefer an invalid intermediate solution, whereas CPL models a conditional selection process that can represent coherent mode commitment.



\end{document}